\documentclass[10pt,twocolumn,letterpaper]{article}

\usepackage{wacv}              % To produce the CAMERA-READY version
\newcommand{\xmark}{\ding{53}}%

\usepackage{graphicx}
\usepackage{amsmath}
\usepackage{amssymb}
\usepackage{pifont}
\usepackage{booktabs}

\definecolor{wacvblue}{rgb}{0.21,0.49,0.74}
\usepackage[pagebackref,breaklinks,colorlinks,allcolors=wacvblue]{hyperref}

\usepackage{multirow}
\usepackage{booktabs}
\usepackage{float}
\usepackage{svg}
\usepackage[normalem]{ulem}
\usepackage{arydshln}

\usepackage[capitalize]{cleveref}
\crefname{section}{Sec.}{Secs.}
\Crefname{section}{Section}{Sections}
\Crefname{table}{Table}{Tables}
\crefname{table}{Tab.}{Tabs.}

\def\wacvPaperID{668} % *** Enter the WACV Paper ID here
\def\confName{WACV}
\def\confYear{2026}

\begin{document}

%%%%%%%%% TITLE - PLEASE UPDATE
% \title{\LaTeX\ Author Guidelines for \confName~Proceedings}
\title{Beyond Paired Data: Self-Supervised UAV Geo-Localization from Reference Imagery Alone}

% \author{First Author\\
% Institution1\\
% Institution1 address\\
% {\tt\small firstauthor@i1.org}
% % For a paper whose authors are all at the same institution,
% % omit the following lines up until the closing ``}''.
% % Additional authors and addresses can be added with ``\and'',
% % just like the second author.
% % To save space, use either the email address or home page, not both
% \and
% Second Author\\
% Institution2\\
% First line of institution2 address\\
% {\tt\small secondauthor@i2.org}
% }

\author{
{
Tristan Amadei$^{1,2}$ \hspace{0.25em}
Enric Meinhardt-Llopis$^{2}$ \hspace{0.25em}
Benedicte Bascle$^{1}$ \hspace{0.25em}
Corentin Abgrall$^{1}$ \hspace{0.25em}
Gabriele Facciolo$^{2}$
} \\[1em]
$^{1}$\small{Thales LAS, France} \\
$^{2}$\small{Universite Paris-Saclay, ENS Paris-Saclay, CNRS, Centre Borelli, France} \\
\small{\texttt{\datasetUrl}} 
}

\maketitle
\begin{abstract}
Image-based localization in GNSS-denied environments is critical for UAV autonomy. Existing state-of-the-art approaches rely on matching UAV images to geo-referenced satellite images; however, they typically require large-scale, paired UAV–satellite datasets for training. Such data are costly to acquire and often unavailable, limiting their applicability.
To address this challenge, we adopt a training paradigm that removes the need for UAV imagery during training by learning directly from satellite-view reference images. This is achieved through a dedicated augmentation strategy that simulates the visual domain shift between satellite and real-world UAV views.
We introduce \textbf{\method}, an efficient model designed to exploit this paradigm, and validate it on \textbf{\dataset}, a new and challenging dataset of real-world UAV images that we release to the community. Our method achieves competitive performance compared to approaches trained with paired data, demonstrating its effectiveness and strong generalization capabilities.
\end{abstract}
    
\section{Introduction}
\label{sec:intro}

Vision-based localization is critical for Unmanned Aerial Vehicles (UAVs) when Global Navigation Satellite Systems (GNSS) are compromised by signal loss or jamming. While techniques like Visual Odometry~\cite{guizilini11} suffer from cumulative drift and Visual SLAM~\cite{blosch10} is often limited to small-scale environments, matching UAV images against a pre-existing database of geo-referenced imagery offers a robust path to absolute positioning.

Approaches to this problem can be broadly categorized. Feature-based matching methods, from traditional SIFT~\cite{lowe99} to modern deep learning models~\cite{sun2021loftr,adamatcher,yu23}, often face computational bottlenecks when scaling to large databases. Consequently, image embedding techniques, which encode images into compact descriptors for efficient comparison, have become the dominant paradigm~\cite{anyloc, earthloc, salad, selavpr, ali2023mixvpr, lu2024cricavpr}. However, these state-of-the-art methods share a critical dependency: they require large-scale, paired training data where query images are matched with reference images. This data requirement presents a major bottleneck for deployment, as collecting extensive, geo-registered UAV imagery for every new flight scenario is often infeasible.

\begin{figure}
    \centering
    
        \setlength{\tabcolsep}{0pt}
        \renewcommand{\arraystretch}{-1}
    
    \begin{tabular}{c} 

        \includegraphics[width=0.45\textwidth,]{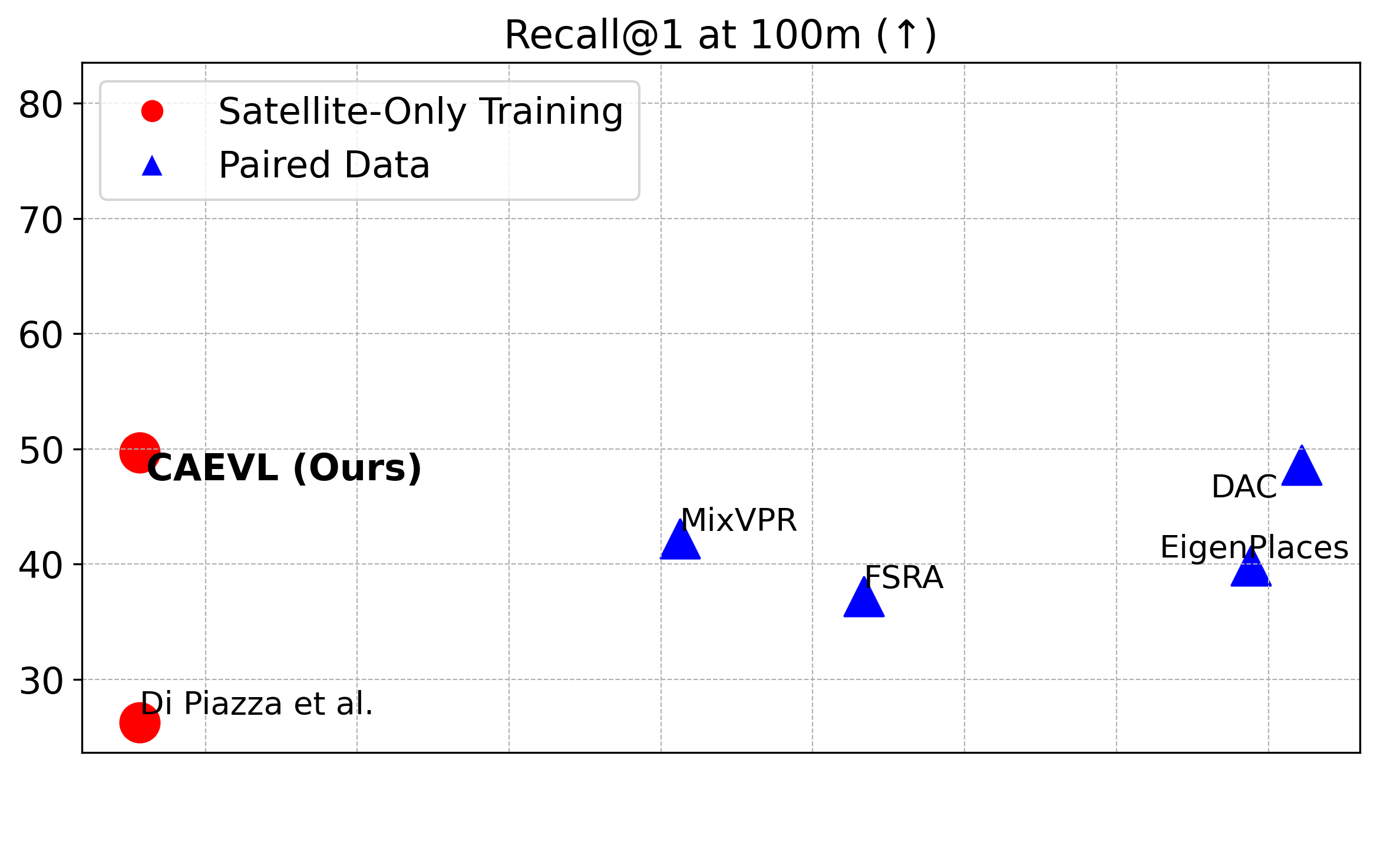} \\
        \includegraphics[width=0.45\textwidth]{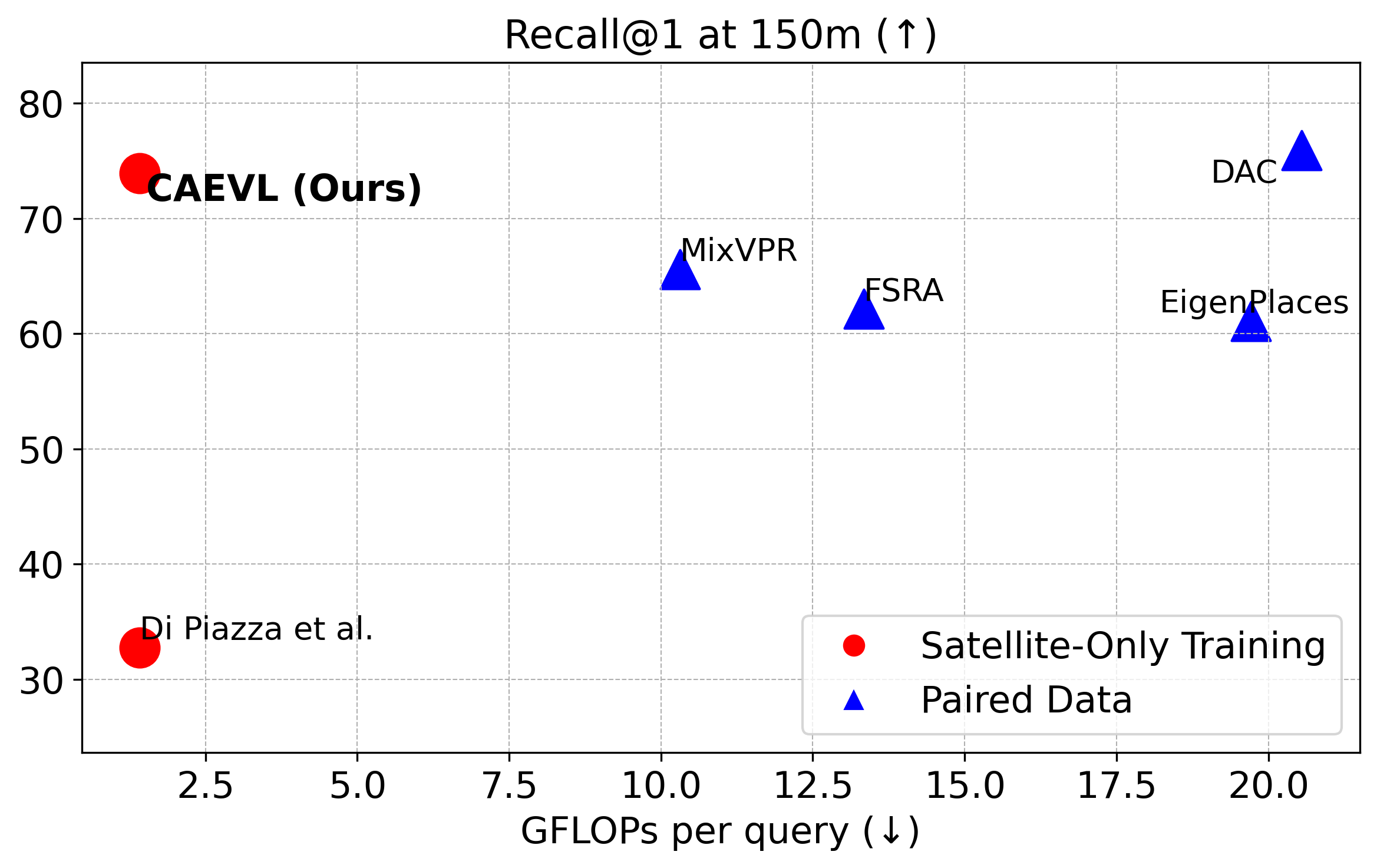}  
        % \\  GFLOPs per query ($\downarrow$)

        % \bottomrule
    \end{tabular}
    \caption{Recall@1 at 100m and 150m for \method and other SOTA methods on the \dataset dataset. \method achieves high accuracy while requiring fewer GFLOPs per query and without using labeled paired data during training. A more comprehensive set of results can be found in \Cref{tab:recall_comparison}.}
    \label{fig: results_comparison}
\end{figure}

Lighter alternatives, such as training an autoencoder on reference images of a specific flight area~\cite{bianchi21, dipiazza2024leveragingedgedetectionneural}, have been proposed but have only been demonstrated on small-scale, ground-level scenes. Furthermore, a simple pixel-wise reconstruction loss is insufficient to create a latent space robust enough to bridge the significant visual discrepancies between satellite and real-world UAV imagery.

To address these challenges, we introduce a novel, data-efficient training paradigm that learns to localize from reference images exclusively, eliminating the need for any paired UAV data during training (\Cref{fig: results_comparison}). Our approach structures the autoencoder's latent space to be resilient to the cross-domain visual shift. This is achieved through two key technical contributions: the integration of a perceptual loss~\cite{johnson2016perceptual} during initial training, and a subsequent non-contrastive fine-tuning stage that regularizes the embedding space to align semantically similar views.

To validate our approach, we introduce \dataset, a new and challenging dataset of real-world UAV images captured during high-altitude flights, which presents conditions such as vignetting and non-nadir orientations that are absent in many existing benchmarks (see \Cref{fig: ign_uav_comparison}). Our main contributions are:
\begin{enumerate}
\item We introduce a data-efficient training paradigm for UAV localization that requires \textbf{only satellite-view reference images}, eliminating the need for paired UAV training data.
\item We present \textbf{\method}, an efficient and lightweight model architecture designed to implement this paradigm, which combines a perceptual loss with a non-contrastive regularization of the latent space.
\item We release \textbf{\dataset}, a new large-scale dataset of real-world UAV images captured under challenging, high-altitude conditions, to validate our approach and facilitate future research in this area. 
% The dataset is available at \datasetUrl.
\end{enumerate}

\section{Related Work}
\label{sec:related_work}

\textbf{Cross-View Geo-Localization Paradigms.}
Visual Geo-Localization aims to find a camera's position by matching a query image against a geo-referenced database. The task becomes Cross-View Geo-Localization (CVGL) when matching images from different perspectives, such as a UAV against a satellite map~\cite{drones8110622, image_object_geoloc, durgam2024cross}. This is inherently challenging due to a domain gap caused by viewpoint, scale, and temporal variations. Two paradigms dominate the field: image retrieval via metric learning, and direct matching for fine-grained localization.

\smallskip\noindent\textbf{Image Retrieval-Based Geo-Localization.}
The most common approach treats CVGL as an image retrieval task, learning a shared embedding space where feature distance corresponds to geographic proximity~\cite{liu2023view, li2025recognition}.

\textit{Ground-Level and General VPR.}  
Early work in Visual Place Recognition (VPR) and ground-to-satellite matching focused on global descriptors, with NetVLAD~\cite{arandjelovic2016netvlad} introducing trainable aggregation layers. Later methods improved robustness to viewpoint changes, such as EigenPlaces~\cite{eigenplaces}, which clusters training data to embed viewpoint invariance, and MixVPR~\cite{ali2023mixvpr}, which mixes features via MLPs for compact descriptors. However, these approaches are tailored to ground-level imagery and do not fully address the top-down perspective of UAVs.

\textit{UAV-to-Satellite Retrieval.}  
In UAV-to-satellite matching, research has evolved from CNNs to Transformers. While CNN-based methods like MCCG~\cite{shen2023mccg} remain relevant, Vision Transformers (ViTs)~\cite{dosovitskiy2020image} dominate due to their ability to model global dependencies and reduce cross-view variance~\cite{huang2024cv, dai2021transformer}. Some works leverage adversarial learning to produce view-invariant features~\cite{liu2023view}, while others address real-world challenges. For instance, AGEN~\cite{agen} combines a DINOv2~\cite{oquab2024dinov2} backbone with a Local Pattern Network~\cite{lpn} and an Adaptive Error Control module to maintain accuracy under extreme conditions.

\smallskip\noindent\textbf{Direct Matching and Fine-Grained Localization.}
An alternative paradigm formulates geo-localization as direct matching or heatmap regression~\cite{yao2024uav, cheng2024offset}. Early GAN-based methods tackled modality gaps via generative translation: LocGAN~\cite{locgan} converts UAV images to LiDAR grids for pose estimation, while \cite{schleiss2019translating} translates UAV imagery into street-map–like views for accurate GNSS-denied localization. Recent supervised GANs, such as SAM-GAN~\cite{xu2023sam} and~\cite{si2024gan}, enhance translation through style/content disentanglement, topological consistency, and edge-aware generation. PVDA~\cite{liu2024viewdistributionalignmentprogressive} instead aligns UAV and satellite features using progressive adversarial learning.

Non-GAN alternatives include BRM localization~\cite{choi2020brm}, that matches building area ratios to maps, and orthophoto-based Monte Carlo localization for probabilistic GNSS-free matching~\cite{kinnari2021gnss}. Transformer-based models like MLPCAN~\cite{li2024aerial}, SSPT~\cite{sspt}, and VRLM~\cite{yao2024uav} leverage cross-attention and multi-scale fusion to produce precise location heatmaps. These are often paired with robust matchers: learned methods such as LoFTR~\cite{sun2021loftr} and SuperPoint/SuperGlue~\cite{detone2018superpoint, sarlin2020superglue} achieve strong performance~\cite{liu24multisource, liu23realtime, uav_mountain_loftr}, though their reliance on repeatable keypoints limits standalone use in extreme cross-view settings.

\smallskip\noindent\textbf{Emerging Trends and Learning Strategies.}
To improve reliability, recent research moves beyond single-image queries, using either unordered sets of images (Set-CVGL~\cite{wu2024cross}) or ordered video sequences~\cite{zhang2023cross} to aggregate spatial and temporal information. The training of these models relies heavily on self-supervision. While contrastive methods~\cite{contrastive_survey, simclr, moco} with triplet loss~\cite{hoffer2015deep} are common, they depend on careful negative sampling~\cite{complexuav}. Non-contrastive approaches like BYOL~\cite{byol20} and VICReg~\cite{vicreg} avoid this by learning from positive pairs alone. Our approach builds on these ideas, integrating autoencoder-based embeddings with non-contrastive learning to better structure the latent space for robust high-altitude cross-view localization.

\smallskip\noindent\textbf{UAV Localization Datasets.}
Benchmark datasets are critical drivers of progress. Foundational datasets like University-1652~\cite{zheng2020university} and SUES-200~\cite{sues} have been largely superseded by more challenging modern benchmarks. Datasets such as DenseUAV~\cite{denseuav}, UAV-VisLoc~\cite{xu2024uav}, AnyVisLoc~\cite{ye2025exploring}, and ComplexUAV~\cite{complexuav} have introduced larger maps, more diverse terrains, and higher altitude challenges. To mitigate the cost of real-world data collection, synthetic datasets like VDUAV~\cite{yao2024uav} and GTA-UAV~\cite{gtauav} have also been proposed. Our \dataset dataset differs significantly, featuring real UAV images from much higher altitudes (up to 1600m) with sensor distortions and lighting variations that are representative of challenging, real-world deployment scenarios.

% \section{\dataset: real-world localization dataset}
\section{\dataset: a new benchmark for high-altitude, real-world challenges}
Visual Localization Dataset (\dataset) is composed of: reference images that are extracted from public aerial image databases over an area of interest; and test data consisting of real-world UAV images captured during real flights, performed by an ultralight aircraft following typical trajectories, camera, and IMU (Inertial Measurement Unit) of a drone. 

UAV images are used only during testing, never during training. This ensures that our method remains generalizable, making it suitable for flying over areas where the drone has never been before and where UAV images are unavailable in advance. Reference images, however, are used both for training and for localizing UAV images during testing.

\begin{figure}
    \centering
    \includegraphics[width=1\linewidth]{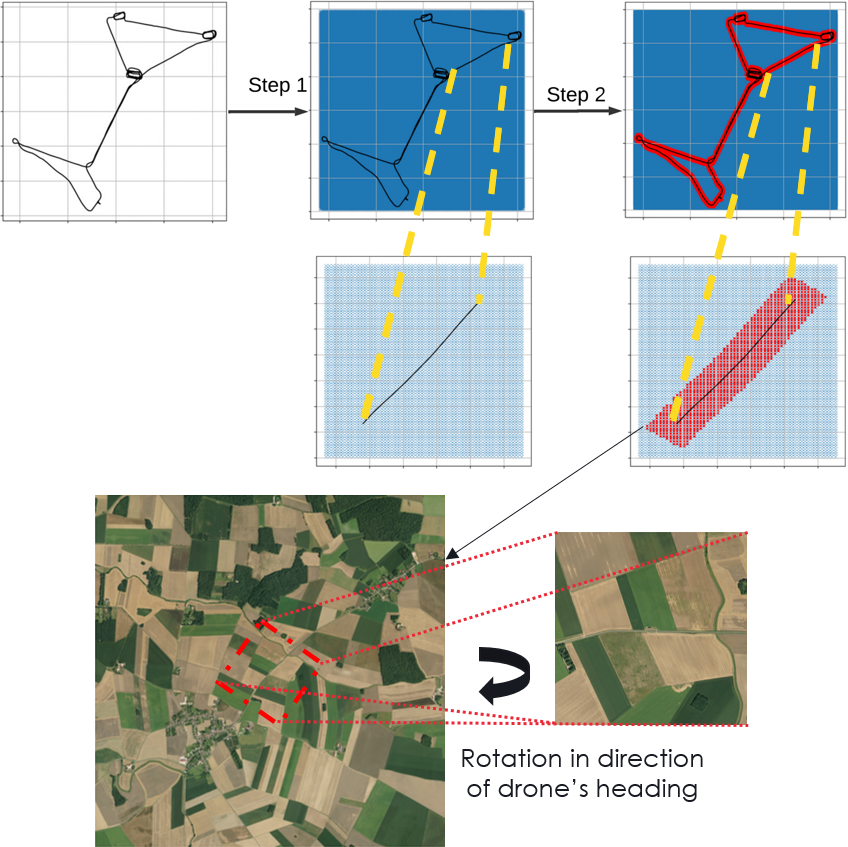}
    \caption{Step 1: define a regular tiling (blue points) that covers the entire flight trajectory. Step 2: define a search zone around the flight trajectory and select the points from the regular tiling inside the search zone (red points). Figures on the lower row show zooms of the upper images, to display closer details. For each geographic point previously selected, we extract a reference image centered around it and rotate it in the direction of the heading of the drone.}
    \label{fig: ign_data_selection_steps}
\end{figure}

\subsection{UAV images}
\label{subsec: uav_images}
The UAV images were recorded during UAV flights. During the flights, the UAV’s altitude, as well as its pitch, roll and heading are precisely tracked. We also know the GPS coordinates of the drone at all times during the flights, which will be used as ground truths during the later tests.
Raw UAV images are distorted due to a fisheye effect on the onboard camera. The images we use in this study are thus processed to correct these distortions. 
We primarily focused on two flights during our study: we have 30,909 UAV images from the first flight, and 59,438 images from the second one.  We will relate the results on the first flight (see \Cref{fig: ign_data_selection_steps}) in this paper. The first flight lasted 5,870 seconds, during which the drone traveled around 75km. It had an average altitude of 571m, with a standard deviation of 116m. \Cref{tab:land_cover_flight1} in the supplementary material shows the average distribution of land cover types in reference data along the first flight. The second flight lasted 11,866 seconds, with a length of 320km. The drone had an average altitude of 1007m, with a standard deviation of 402m. Further details of both flights are to be found in \Cref{sec:dataset_details} of the supplementary material.

\subsection{Reference data}
The IGN DB Ortho \cite{ign} database was used to extract the reference aerial images for the localization task. This database consists of a collection of georeferenced orthophotographs covering the entire French territory, typically produced at a 20 cm resolution, in both RGB and near-infrared formats. The database is updated every 3-4 years, thus for each location, multiple images from different years are available. Note that the IGN DB Ortho images have spectral characteristics and resolution comparable to those of very high-resolution satellite images.

To construct the reference database for the localization of a flight, we define a 1400m zone wrapping the flight trajectory (700m on each side), as illustrated on \Cref{fig: ign_data_selection_steps}. We then build a regular tiling covering the whole flight trajectory, and extract reference images from IGN tiles at coordinates both within the tiling and the 1400m band. 
The images are cropped to simulate the drone’s vision, taking into account the camera’s focal length, average flight altitude, and planned flight path, all of which are known in advance. From this information, the appropriate crop size for the IGN images can be determined. Reference images are also rotated to align with the drone’s planned heading, which is available at all times via the drone’s onboard instruments. 

\Cref{fig: ign_data_selection_steps} shows the steps to extract an image from a point within the search zone.
We end up with 225,290 reference images for our first flight, and 401,773 for the second flight.

\begin{figure}
    \centering
    \includegraphics[width=1\linewidth]{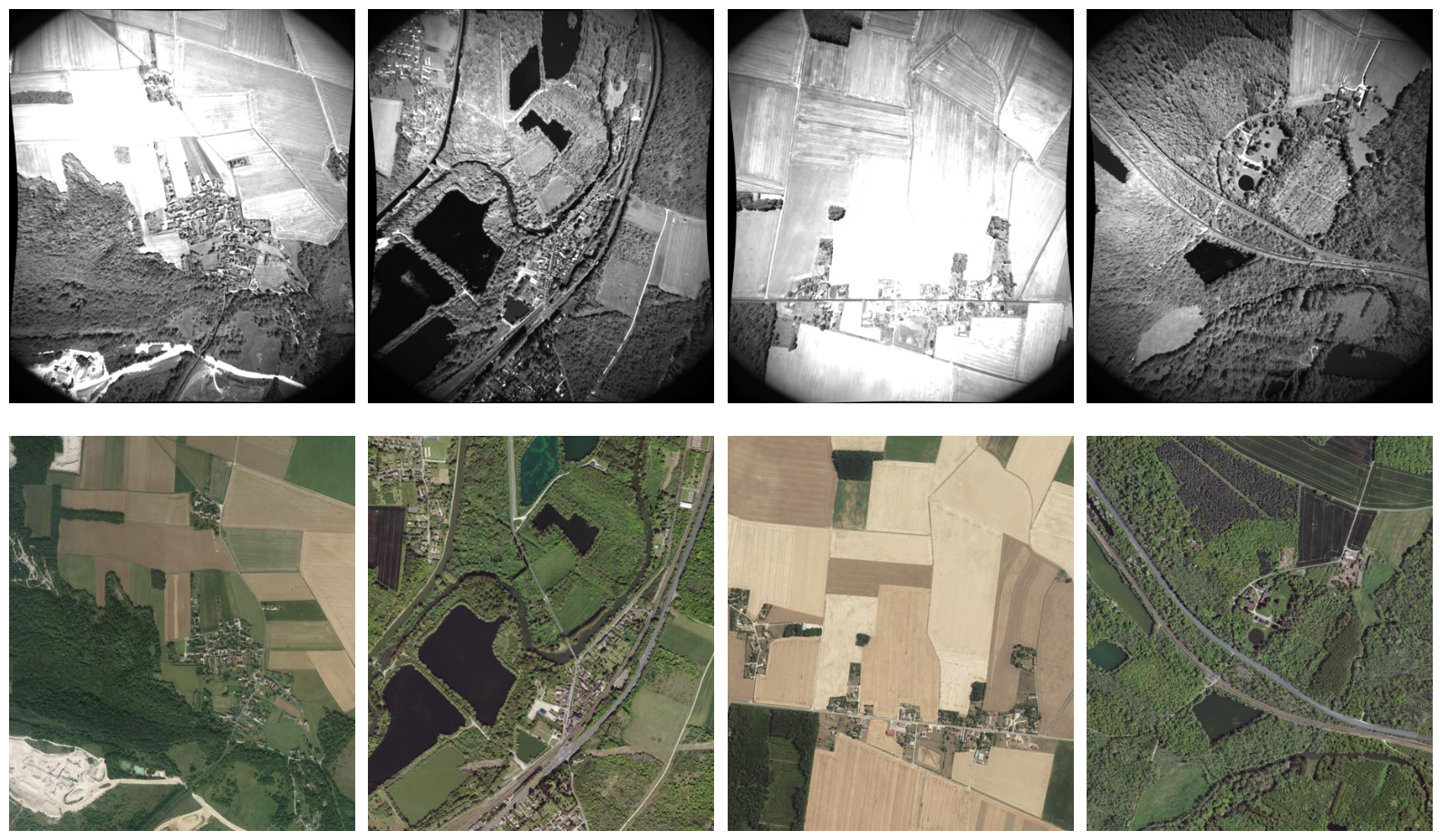}
    \caption{Examples of randomly picked UAV images (upper row) and their geographically closest reference images (lower row). These images are extracted from our proposed dataset \dataset.}
    \label{fig: ign_uav_comparison}
\end{figure}

\subsection{UAV and reference images differences}
\label{subsec: uav_ign_diffs}
Significant visual differences exist between UAV and reference images, as illustrated in \Cref{fig: ign_uav_comparison}. A key difference is vignetting. In addition, unlike reference images, the UAV images are not consistently  captured with a nadir orientation. As the images are assumed to be at nadir, these imprecisions induce some  alignment errors with the ground.

Another differentiating factor is altitude. We assume the average altitude of the UAV throughout the flight is known, but not its precise altitude at each timestamp, due to inherent inaccuracies in onboard altitude estimation. Consequently,  while reference images are extracted from the IGN basemap at a fixed altitude based on the UAV’s average flight altitude, individual UAV images may reflect some altitude discrepancies relative to their corresponding references. 

\subsection{Differences with existing UAV datasets}
\label{subsec:diff_datasets}
Datasets with aerial UAV images for cross-view geo-localization do exist in the literature, such as \cite{sues, he2023foundloc, schleiss2022vpair, alto, complexuav, denseuav}. 
However, our \dataset dataset differs significantly from existing ones in several ways. Notably, \dataset includes images taken by drones at much higher altitudes (up to 1600m, compared to 300-400m for other datasets). \dataset features images captured under varying lighting conditions, with distortions and viewpoints that are not always at nadir, making it more challenging, reflecting real-world operational conditions. Lastly, available datasets contain images with many visual features, while the UAV images in our dataset often contain redundant information and few distinctive visual features, making them much harder to localize. As shown in \Cref{tab:land_cover_flight1} in the supplementary material, the high-altitude images in \dataset mostly capture fields and forests, with few distinguishing visual features. In contrast, existing datasets contain images taken at low or medium altitudes, where each image includes more distinctive structures and visual details.
Figures \ref{fig:vpr_datasets_comparison} and \ref{fig:vpr_datasets_comparison_1} in the supplementary material show a random query image and its corresponding reference image from \dataset and from other aerial visual place recognition datasets.

% \begin{figure}
%     \centering
%     \includegraphics[width=1\linewidth]{images/dataset_comparison/0.png}
%     \caption{Comparison of a randomly selected query-reference pair for different VPR datasets.}
%     \label{fig:vpr_datasets_comparison}
% \end{figure}

\begin{figure*}
    \centering
    \includegraphics[width=1\linewidth]{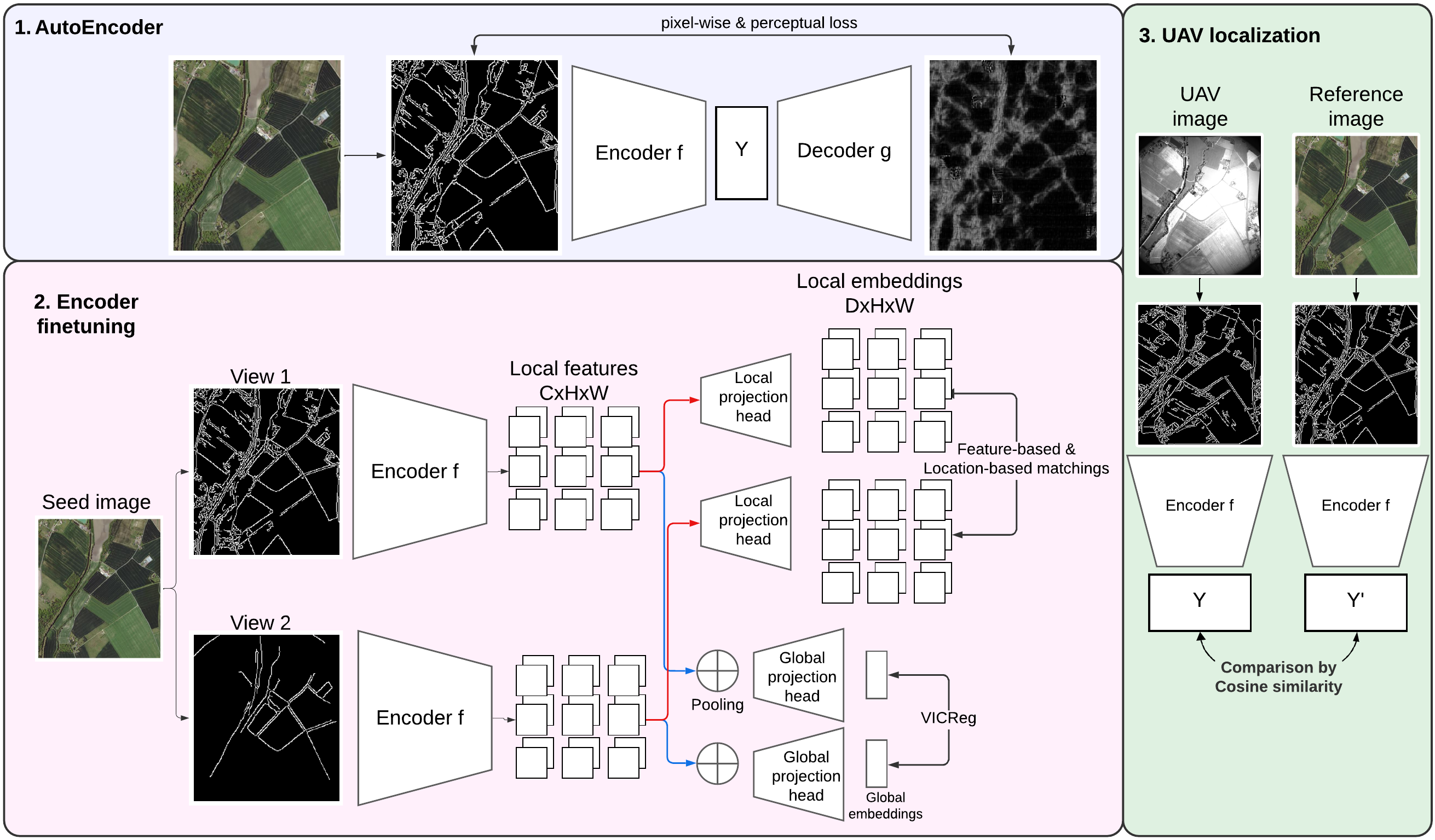}
    % \includesvg[width=1\linewidth]{images/model_overview_3steps.svg}
    \caption{Overview of \method. All input images are first processed through a Canny filter to extract the edges. An autoencoder is trained using a pixel-wise $L_2$ loss and a perceptual loss. The decoder is then discarded and the encoder is fine-tuned using a non-contrastive approach. Two views of the same input image are passed through the encoder to produce both local features (feature maps before pooling) and global features (embeddings after pooling). The local features are fed to a local projection head to project them to a smaller space. Two sets of matches are computed: one using the spatial information from each view, and the other based on the $L_2$-distance in the embedding space. The VICReg criterion is then applied to these matched spatial embeddings. Furthermore, the global features are passed into a global projection head to produce global embeddings. The VICReg criterion is applied on these global embeddings. After training, the local and global projection heads are discarded. The encoder is kept to compute embeddings of UAV and reference images, that will be compared using the cosine similarity.}
    \label{fig: overview}
\end{figure*}

\section{Method}
\label{sec: methodology}

Our method, \method, follows a two-stage training process, as illustrated in \Cref{fig: overview}. Following prior work~\cite{dipiazza2024leveragingedgedetectionneural, afolabi15}, all images are first pre-processed with a Canny filter to extract structural edges, a choice validated in our ablation study (\Cref{subsec: ablation}). An autoencoder is then pre-trained on these edge maps using a combination of pixel-wise $L_2$ and perceptual losses. Subsequently, the decoder is discarded, and the encoder is fine-tuned with a non-contrastive, self-supervised objective to learn domain-invariant features. For inference, the final encoder generates embeddings for query and reference images, which are then matched using cosine similarity.

\subsection{Perceptual loss}
To overcome the limitations of pixel-wise losses, which often fail to capture perceptually important details, we incorporate a perceptual loss~\cite{johnson2016perceptual}. This loss computes discrepancies in a high-level feature space rather than at the pixel level, encouraging the model to preserve structural content. The combined autoencoder loss is:
\begin{equation}
L_{AE} = ||I - D(E(I))||_2^2 + \beta \cdot \sum_l ||\varphi_l(I) -\varphi_l(D(E(I)))||_2^2,
\label{eq: perceptual}
\end{equation}
where $I$ is the input image, $D$ and $E$ represent the encoder and decoder respectively, and $\varphi_l$ refers to the feature map at layer $l$ of a pre-trained backbone. 

For our task, which requires robustness to the visual discrepancies between UAV and satellite imagery, we leverage the powerful features from a pre-trained DinoV2~\cite{oquab2024dinov2} backbone to compute $\varphi_l$.

% \subsection{Latent space regularization}
\subsection{Simulating the cross-view domain shift}
\label{subsec: non-contrastive}

Simple image reconstruction is insufficient for this task, as the model must learn to bridge the domain gap between satellite and UAV imagery without seeing paired examples. To achieve this, we fine-tune the encoder using a non-contrastive, self-supervised objective. The core of this stage is an augmentation pipeline that transforms a single satellite image into two distinct views, which are then treated as a positive pair. This pipeline includes standard transformations (e.g., rotation, blur) and, critically, specific augmentations like vignetting to explicitly simulate the visual artifacts of real-world UAV data.

By training the encoder to produce invariant features for these augmented pairs, the model learns to bridge the domain gap. We employ the VICRegL framework~\cite{bardes2022vicreglselfsupervisedlearninglocal} for this fine-tuning stage, as its multi-scale approach applies the VICReg criterion~\cite{vicreg} to both global embeddings and local feature maps. This enforces spatial consistency and prevents informational collapse, making it highly effective for our localization task.

We now detail the matching strategies applied to the local feature maps.

\smallskip\noindent\textbf{Location-based matching} to account for the transformation between views $x$ and $x'$ of an image $I$. After processing $x$ and $x'$ in the encoder and the local projection head, we get the local embeddings $z$ and $z'$, of dimension $D \times H \times W$. For each feature vector $z_p$ at position $p$ in the local embedding (i.e. $z_p \in \mathbb{R}^D$), we find its nearest spatial neighbor $z'_{\text{NN}(p)}$ in $z'$ based on their absolute positions in $I$. Among the resulting $H \times W$ pairs, only the top-$\gamma$ pairs are retained. The location-based matching loss function is defined as
\begin{equation}
\label{eq: loc_based}
L_s(z, z') = \sum_{p \in P} l(z_p, z'_{\text{NN}(p)}),
\end{equation}
where $P = \{(h, w) \,|\, (h, w) \in [1, \dots, H] \times [1, \dots, W]\}$ represents all coordinates in the feature map, and $\text{NN}(p)$ denotes the spatially nearest coordinate to $p$ based on their positions in the original image $I$.

\begin{table*}
\centering
\resizebox{0.99\textwidth}{!}{%
\begin{tabular}{@{}l c ccc ccc ccc ccc c c@{}}
\toprule
& \multirow{2}{*}{\shortstack{\textbf{Satellite-Only} \\ \textbf{Training}}}  & \multicolumn{3}{c}{\textbf{100m}} & \multicolumn{3}{c}{\textbf{150m}} & \multicolumn{3}{c}{\textbf{250m}} & \multicolumn{3}{c}{\textbf{500m}} & \multirow{2}{*}{\shortstack{\textbf{Descriptors} \\ \textbf{Dimension}}} & \multirow{2}{*}{\shortstack{\textbf{GFLOPs} \\ \textbf{per query}}}\\
\cmidrule(lr){3-5} \cmidrule(lr){6-8} \cmidrule(lr){9-11} \cmidrule(lr){12-14}
Method & & R@1 & R@5 & R@10 & R@1 & R@5 & R@10 & R@1 & R@5 & R@10 & R@1 & R@5 & R@10 &  & \\ 
\midrule
\midrule
Random & - & 0.03 & 0.16 & 0.33 & 0.07 & 0.36 & 0.73 & 0.21 & 1.02 & 2.03 & 0.86 & 4.16 & 8.06 & - & - \\
\midrule
SuperPoint~\cite{detone2018superpoint}~+~LightGlue~\cite{lightglue}* & \xmark & 11.60 & 33.80 & 48.40 & 24.80 & 54.00 & 67.40 & 49.40 & 78.20 & 86.20 & 86.20 & 95.60 & 97.60 & 256 & 17,705.44 \\
\midrule

MixVPR~\cite{ali2023mixvpr} (Zero-shot) & \multirow{2}{*}{\xmark} & 14.16 & 37.60 & 49.62 & 27.60 & 54.15 & 63.98 & 54.12 & 72.27 & 77.09 & 76.08 & 81.40 & 82.98 & \multirow{2}{*}{4096} & \multirow{2}{*}{\underline{10.31}} \\
MixVPR (Finetuned) & & 42.19 & 66.44 & 73.96 & 65.58 & 78.26 & 81.95 & 79.22 & 83.28 & 84.41 & 83.12 & 85.44 & 86.44\\ [2mm]

EigenPlaces~\cite{eigenplaces} (Zero-shot) & \multirow{2}{*}{\xmark} & 11.43 & 28.81 & 38.22 & 22.43 & 41.49 & 50.30 & 39.81 & 55.93 & 62.29 & 58.50 & 67.50 & 71.37 & \multirow{2}{*}{2048} & \multirow{2}{*}{19.71} \\
EigenPlaces (Finetuned) & & 39.84 & 66.53 & 74.37 & 61.09 & 77.91 & 81.45 & 78.71 & 83.35 & 84.58 & 82.01 & 84.62 & 85.67 \\ [2mm]

Megaloc~\cite{berton_2025_megaloc} (Zero-shot) & \xmark & 3.07 & 10.66 & 16.28 & 7.15 & 19.59 & 26.98 & 18.63 & 36.36 & 44.72 & 41.69 & 57.79 & 63.70 & 8448 & 54.93\\
\midrule
FSRA~\cite{dai2021transformer} (Zero-shot) & \multirow{2}{*}{\xmark} & 1.06 & 3.96 & 6.28 & 2.26 & 6.91 & 10.12 & 5.71 & 12.95 & 17.02 & 14.39 & 24.61 & 34.11 & \multirow{2}{*}{\textbf{512}} & \multirow{2}{*}{13.34} \\
FSRA (Finetuned) & & 37.19 & 68.41 & 78.50 & 62.13 & 83.28 & \underline{87.80} & \underline{84.69} & \underline{90.46} & \underline{91.92} & \underline{91.37} & \underline{93.40} & \underline{94.03} \\ [2mm]
DAC~\cite{xia2024enhancing} (Zero-shot) & \multirow{2}{*}{\xmark} & 0.76 & 3.20 & 5.27 & 2.20 & 6.39 & 9.10 & 5.56 & 11.69 & 14.77 & 13.80 & 22.87 & 27.05 & \multirow{2}{*}{\underline{1024}} & \multirow{2}{*}{20.55} \\
DAC (Finetuned) & & \underline{48.58} & \textbf{74.67} & \textbf{80.33} & \textbf{75.92} & \textbf{85.67} & \textbf{89.50} & \textbf{85.47} & \textbf{91.14} & \textbf{93.26} & \textbf{93.25} & \textbf{95.95} & \textbf{96.42} \\

\midrule

Di~Piazza~\etal~\cite{dipiazza2024leveragingedgedetectionneural} (Finetuned) & \checkmark & 26.26 & 35.71 & 39.47 & 32.77 & 41.62 & 45.56 & 34.09 & 43.58 & 47.85 & 34.79 & 44.81 & 49.43 & \underline{1024} & \textbf{1.42} \\ 
\midrule
\textbf{\method \space (Ours)} & \checkmark & \textbf{49.68} & \underline{72.80} & \underline{78.87} & \underline{73.93} & \underline{82.32} & {83.96} & {82.80} & {84.50} & {85.83} & {84.16} & {85.36} & {85.96} & \underline{1024} & \textbf{1.42} \\
\bottomrule
\end{tabular}%
}
\caption{Comparison of Recall@K performance for various methods at different localization thresholds (100m, 150m, 250m, and 500m) on the \dataset  dataset. Each method's descriptor dimension and computational cost (GFLOPs) per image query are also reported. \textbf{Bold} values indicate the best in each column, while \underline{underlined} values represent the second best.
*For SuperPoint and LightGlue, feature extraction and matching between a single query and reference image takes an average of 0.012s. Matching one query image to all reference images would take too long. To reduce computational time, we selected 500 query images from the test set and matched each to reference images within a 1km radius. For a fair comparison, we reproduce this experiment with \method \space and the other methods in the supplementary material.}
\label{tab:recall_comparison}
\end{table*}

\smallskip\noindent\textbf{Feature-based matching} to capture long-range interactions. In this approach, each feature vector $z_p$ at position $p$ in the original image is matched to its nearest neighbor $\text{NN}(z_p)$ in the embedding space $z'$ based on the $L_2$-distance. Among the resulting $H \times W$ pairs, only the top-$\gamma$ pairs are retained. The feature-based matching loss is defined as
\begin{equation}
\label{eq: feature_based}
L_d(z, z') = \sum_{p \in P} l(z_p, \text{NN}(z_p)),
\end{equation}
where $\text{NN}(z_p)$ denotes the closest feature vector to $z_p$ in the feature maps $z'$, with respect to the $L_2$-distance. 
The final loss combines both global and local matching strategies, defined as
\begin{align}
\label{eq: vicregl}
    L(z, z') &= \alpha \cdot \ell(z_*, z'_*) +  \\
     (1-\alpha) &\cdot [L_s(z, z') + L_s(z', z) + L_d(z, z') + L_d(z', z)],\nonumber
\end{align}
where $z_*$ is the embedding after pooling, $z$ is the embedding before pooling, and $\ell$ is the VICReg criterion.

\smallskip\noindent\textbf{Image pair generation}. To introduce variability between UAV and reference images, specific augmentations were selected to simulate condition differences. This included vignetting, reflecting image discrepancies between the two modalities, as well as translation, rotation, zoom, brightness/contrast adjustments, Gaussian noise, and blur. Implementation details for the augmentations are given in the supplementary material. This augmentation strategy helped enforce a diverse data environment during training. These augmentations transform a reference image to resemble a UAV image, signaling to the model that these distinct images should be closely aligned in the latent space. They also promote a smooth latent space, ensuring that semantically similar images (\ie two augmented views of the same image) have similar representations. 

\subsection{UAV localization}

To localize a UAV image against a reference database, we compute the embedding of the UAV image and those of the reference images. We then compute the cosine similarity between the UAV embedding and each reference embedding, which act as similarity scores to be maximized. The reference image with the highest similarity score is considered as the predicted image, and we consider the location of its center as the predicted localization of the UAV image.

\section{Experiments, results and analyses}
\label{sec: experiments}

The autoencoder architecture of \method  is inspired by~\cite{hou2017deep} and~\cite{dipiazza2024leveragingedgedetectionneural}. Its encoder comprises 4 strided convolutional layers with batch normalization and \textit{LeakyReLU} activation, leading to a 1024-dimensional latent space. The encoder, which is the only part of the method to be embedded on the UAV, has 33 million parameters. Further implementation details about the model's architecture and the augmentations are to be found in the supplementary material.

For ease of read and conciseness, we will only relate the results of our method on the first flight, with 30,909 images.

\subsection{Performance on \dataset}
\label{subsec: implementation_details}

To provide a thorough and fair analysis on our proposed \dataset dataset, we compare \method against state-of-the-art methods under two distinct training conditions. The first condition represents a data-hungry, fully supervised paradigm where we fine-tune the baseline methods on the full paired training split, which contains both aerial reference images and their corresponding UAV query images. This demonstrates their performance potential when given access to extensive in-domain training data. In contrast, the second condition is data-efficient and aligns with our core contribution. In this paradigm, we train our proposed \method model using only the aerial reference images from the \dataset training split; the model never sees a single UAV image during training. This experimental setup allows us to directly test our central hypothesis that a data-efficient, satellite-only approach can achieve competitive performance compared to data-hungry methods, thus offering a more practical solution for real-world deployment.

We follow the standard place recognition evaluation procedure \cite{arandjelovic2016netvlad, ali2023mixvpr, zaffar2021vpr, salad, earthloc, eigenplaces}, where a query is correctly localized if at least one of the top K retrieved reference images is within d = 100, 150, 250, or 500 meters of the ground truth. Recall@K (K = 1, 5, 10) is used to evaluate localization performance at different localization tolerances. Scores are computed at 100m and beyond, rather than the typical 15m or 25m, to accommodate the higher altitudes in our dataset, where a single pixel covers a larger ground area than in other aerial VPR datasets.

We compare our method, \method, against a suite of state-of-the-art VPR and geo-localization methods in \Cref{tab:recall_comparison}. The baselines were evaluated both in a zero-shot capacity and after being fully fine-tuned on our dataset, providing a comprehensive view of their capabilities.

\begin{figure}[t]
    \centering
    \includegraphics[width=1.\linewidth]{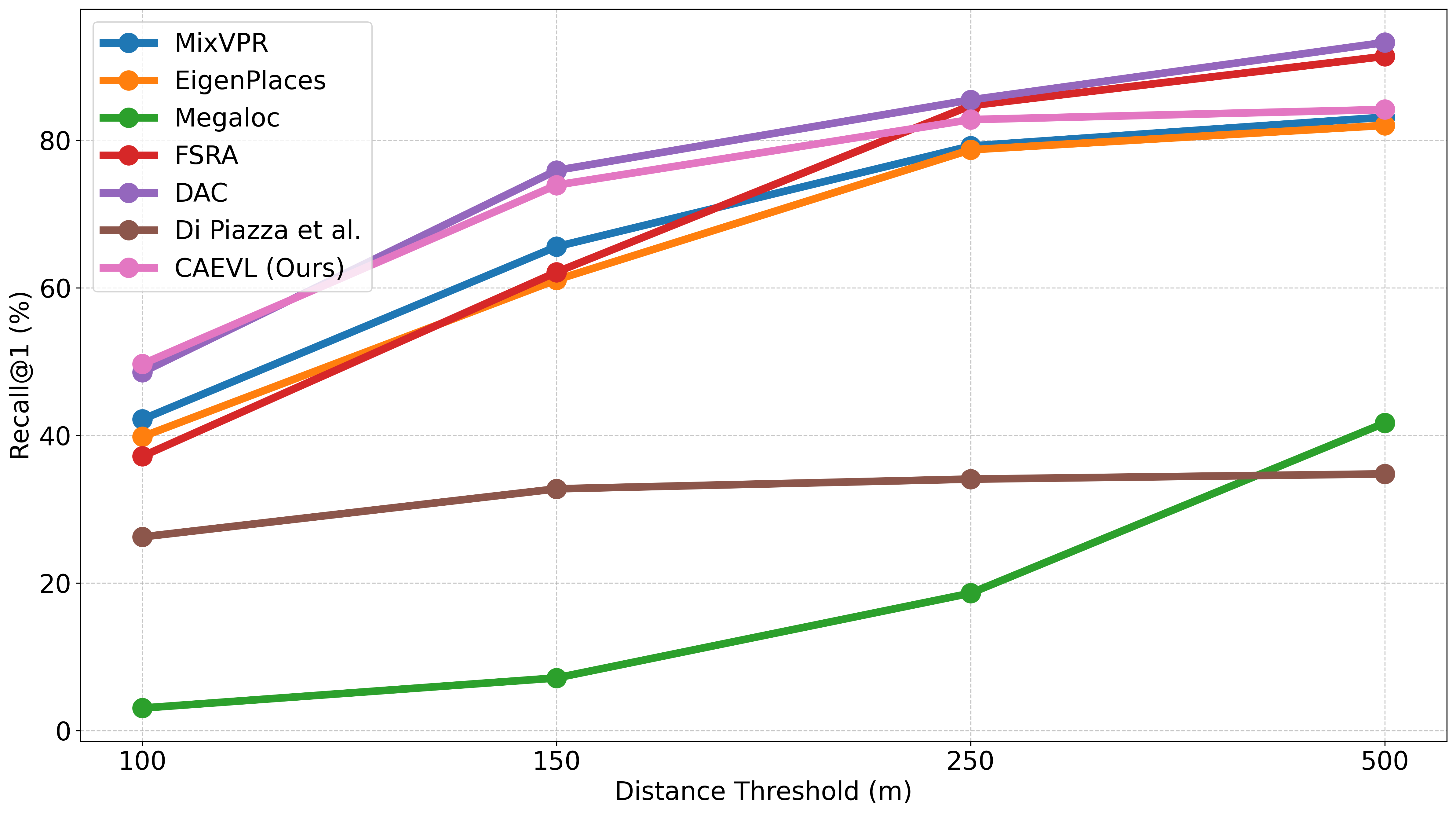}
    \caption{Evolution of Recall@1 across different distance thresholds (100m, 150m, 250m, 500m) for all evaluated methods, using fine-tuned models when applicable. Higher values indicate better localization performance.}
    \label{fig: results_curve@1}
\end{figure}

The evolution of Recall@1 across distance thresholds is shown in \Cref{fig: results_curve@1}. Results highlight the effectiveness of our ``reference-only" training paradigm. The large performance gap between zero-shot and fine-tuned baselines confirms that \dataset is challenging and that existing methods require in-domain adaptation with paired data to be effective. This gap also exposes the limited generalization of pre-trained models on current public benchmarks, emphasizing the value of our dataset.

Our method achieves high accuracy without using any UAV query images during training. While fully fine-tuned methods like FSRA and DAC reach the highest recall at larger thresholds, \method remains highly competitive, especially at stricter 100m and 150m thresholds.

Moreover, this accuracy comes with excellent computational efficiency. At only 1.42 GFLOPs per query, our model is an order of magnitude lighter than other high-performing fine-tuned competitors. Compared to Di Piazza \etal, \method achieves substantially higher accuracy, showing the strength of our training strategy.

All reference images are encoded to produce descriptors, with every query matched against the entire reference database. In practice, reference descriptors are precomputed offline, so inference cost includes only the query encoding and similarity search.

\begin{figure}[t]
    \centering
    \includegraphics[width=1.\linewidth]{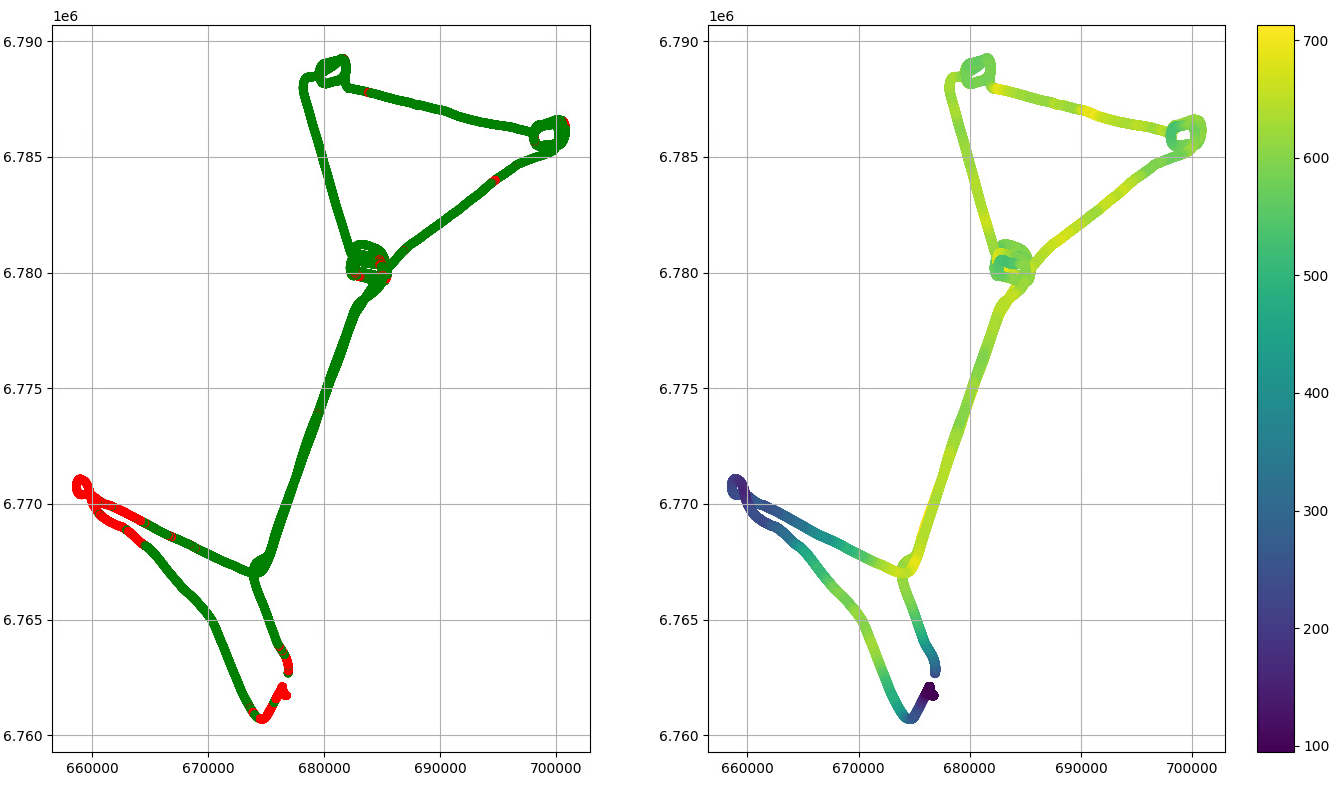}
    \caption{Left: trajectory of flight 1, green points correspond to correct top-1 UAV localizations within 500m and red points correspond to wrong localizations. Right: illustration of the altitude of the drone throughout the flight, in meters.}
    \label{fig: vol10_traj_altitude}
\end{figure}

\Cref{fig: vol10_traj_altitude} highlights the flight sections where \method \space achieved accurate localization within 500m of ground truth on the \dataset \space dataset. Most localization failures generally occur over continuous intervals, often coinciding with significant altitude drops. Although \method \space can handle moderate altitude variations, larger drops challenge the model due to the altered viewpoint. Further experiments are detailed in the supplementary material.

% To assess the temporal consistency of our method, we measured the average length of consecutive queries that were not well predicted. In this study, a query is considered wrongly predicted when its top-1 prediction is farther than 100m from the ground truth. Under this definition, the average consecutive length of wrong predictions is 2.37 frames (and a maximum of 16 wrongly predicted frames in a row), which decreases to 1.82 when using a 150m threshold. If we instead consider a query well predicted when any of the top-5 predictions falls within the distance threshold, the averages become 2.03 (100m) and 1.78 (150m). 
% This indicates that even where errors occurred, they were typically short-lived, showing that our method quickly recovered from incorrect predictions, as we can see on \Cref{fig: vol10_traj_altitude}.

To evaluate temporal consistency, we measured the average length of consecutive frames with incorrect predictions, where a prediction is considered wrong if its top-1 result is more than 100m from the ground truth. Under this criterion, the average run of consecutive errors is 2.37 frames, which decreases to 1.82 with a 150 m threshold. When considering a prediction correct if any of the top-5 results falls within the threshold, these averages drop to 2.03 (100m) and 1.78 (150m). Overall, errors remain short-lived, indicating that the method quickly recovers from incorrect predictions, as illustrated in \Cref{fig: vol10_traj_altitude}.

\subsection{Comparison on other datasets}
\label{subsec:comparison_other_datasets}

To assess the generalization capabilities of our approach beyond our primary high-altitude domain, we evaluated its performance on three diverse public benchmarks. The results are presented in \Cref{tab:recall15m_datasets}.

% \smallskip\noindent\textbf{VPAIR}~\cite{schleiss2022vpair} is a real-world dataset composed of 2,706 query-reference pairs captured by a downward-facing camera on a light aircraft at approximately 300 meters above ground. The 100 km trajectory covers a mix of urban, farmland, and forested areas and includes 10,000 distractor images. To ensure a fair comparison, we rotated each query image to match its corresponding reference heading, aligning VPAIR with \dataset.

% \smallskip\noindent\textbf{ALTO}~\cite{alto} is another real-world dataset from an extensive trajectory flown by a helicopter at over 300 meters. It provides high-precision GPS-INS ground truth and downward-facing RGB imagery.

% \smallskip\noindent\textbf{DenseUAV}~\cite{denseuav} is a large-scale, synthetically generated dataset known for its high density of images and photorealistic urban scenes. It presents a different challenge, testing the model's ability to transfer its learning to simulated environments with perfectly aligned data.

\textbf{VPAIR}~\cite{schleiss2022vpair} is a real-world dataset composed of 2,706 query-reference pairs captured by a downward-facing camera on a light aircraft at approximately 300 meters above ground. The 100 km trajectory covers a mix of urban, farmland, and forested areas and includes 10,000 distractor images. To ensure a fair comparison, we rotated each query image to match its corresponding reference heading, aligning VPAIR with \dataset. \textbf{ALTO}~\cite{alto} is another real-world dataset from an extensive trajectory flown by a helicopter at over 300 meters. It provides high-precision GPS-INS ground truth and downward-facing RGB imagery. \textbf{DenseUAV}~\cite{denseuav} is a large-scale, synthetically generated dataset known for its high density of images and photorealistic urban scenes. It presents a different challenge, testing the model's ability to transfer its learning to simulated environments with perfectly aligned data.

On the real-world VPAIR and ALTO datasets, our method obtains the highest Recall@1 scores (\Cref{tab:recall15m_datasets}), indicating strong performance for precise localization, while remaining competitive with other methods at higher values of K. Conversely, on the synthetic DenseUAV benchmark, methods such as MegaLoc and DAC achieve higher recall scores.
This performance characteristic is consistent with our model's design. \method's reliance on edge-based features is optimized for overcoming real-world visual challenges like lighting and seasonal changes, whereas the dense, clean textures prevalent in synthetic data may better suit methods designed to leverage rich color and texture information. 

\begin{table}
\centering
\resizebox{0.49\textwidth}{!}{%
\begin{tabular}{@{}l ccc ccc ccc@{}}
\toprule
& \multicolumn{3}{c}{\textbf{VPAIR}~\cite{schleiss2022vpair}} & \multicolumn{3}{c}{\textbf{ALTO}~\cite{alto}} & \multicolumn{3}{c}{\textbf{DenseUAV}~\cite{denseuav}}\\
\cmidrule(lr){2-4} \cmidrule(lr){5-7} \cmidrule(lr){8-10}
Method & R@1 & R@5 & R@10 & R@1 & R@5 & R@10 & R@1 & R@5 & R@10 \\
\midrule
\midrule

MixVPR & 48.04 & 71.73 & 77.79 & 45.19 & \textbf{89.85} & \textbf{95.57} & 32.44 & 61.92 & 73.79 \\
EigenPlaces & 45.05 & 72.98 & 79.56 & 39.61 & 77.67 & 91.15 & 16.06 & 31.42 & 41.21 \\
MegaLoc & 27.64 & 66.48 & 73.95 & 35.21 & 75.59 & 91.92 & \textbf{55.42} & \underline{70.19} & \underline{75.70}\\
\midrule
FSRA & 36.92 & 66.67 & 72.84 & 43.68 & {83.02} & 93.53 & 37.95 & 56.51 & 66.14 \\
DAC & \underline{49.87} & \textbf{79.45} & \textbf{86.91} & \underline{50.46} & \underline{85.02} & \underline{95.53} & \underline{46.80} & \textbf{72.57} & \textbf{76.02} \\
\midrule
\textbf{\method \space (Ours)} & \textbf{50.73} & \underline{73.39} & \underline{80.23} & \textbf{52.43} & {82.46} & {94.38} & {36.66} & {62.61} & {70.22} \\
\bottomrule
\end{tabular}%
}
\caption{Comparison of Recall@K (R@1, R@5, R@10) at 15m for VPAIR and ALTO, and 100m for DenseUAV.}
\label{tab:recall15m_datasets}
\end{table}

\subsection{Qualitative analysis}
\Cref{subsec: supp__pred_examples} in the supplementary material provides several examples of top-5 predictions output by \method and other baselines on \dataset and the other datasets we experimented with. Those figures show that \method is on par with the SOTA methods across diverse types of terrain, from rural/agricultural to urban.

% \subsection{Ablation studies}
\section{Ablation studies}
\label{subsec: ablation}

We conducted an ablation study (\Cref{tab: ablation_table}) to validate the contribution of each component of our method. Removing the initial Canny edge extraction (row 2) degrades performance, confirming that it helps the model focus on persistent structures while ignoring transient textural details. 
Edge detection, such as Canny, emphasizes stable structural features like roads, buildings, and terrain boundaries while suppressing transient textures from vegetation, shadows, or seasonal changes. This simplifies the visual input, providing a compact and consistent representation. As a result, the encoder processes less redundant information, improving both learning stability and computational efficiency.

Similarly, training the initial autoencoder without the perceptual loss (row 3) results in lower accuracy, highlighting its importance in creating a meaningfully structured latent space. The subsequent non-contrastive fine-tuning stage (row 4) provides a significant boost by improving the model's robustness to the geometric and photometric variations between UAV and reference imagery. 

\begin{table}
\centering
\resizebox{0.45\textwidth}{!}{%
\begin{tabular}{@{}l ccc ccc ccc@{}}
\toprule
& \multicolumn{3}{c}{\textbf{100m}} & \multicolumn{3}{c}{\textbf{500m}} \\
\cmidrule(lr){2-4} \cmidrule(lr){5-7}
Method & \small R@1 & \small R@5 & \small R@10 & \small R@1 & \small R@5 & \small R@10 \\
\midrule
\textbf{\method} & \textbf{49.68} & \textbf{72.80} & \textbf{78.87} & \textbf{84.16} & \textbf{85.36} & \textbf{85.96} \\
\method \space w/o edge extraction & 28.86 & 39.71 & 43.76 & 43.60 & 56.76 & 62.63 \\
\method \space w/o perceptual loss & 41.52 & 61.02 & 66.47 & 69.73 & 73.60 & 75.67 \\
\method \space w/o VICRegL & 41.07 & 57.13 & 62.01 & 62.76 & 71.96 & 75.66 \\
\method \space w/o cosine similarity & 14.37 & 27.04 & 32.21 & 31.60 & 36.22 & 39.44 \\

\bottomrule
\end{tabular}%
}
\caption{Recall@K comparison of \method \space with and without Canny processed images, perceptual loss, VICRegL and using the scalar product as similarity score rather than the cosine similarity.}
\label{tab: ablation_table}
\end{table}
\section{Conclusion}
\label{sec:conclusion}

In this work, we presented a solution to the challenge of high-altitude, UAV-to-satellite cross-view geo-localization, with a focus on addressing the data-dependency issues of existing methods. We introduced a data-efficient training paradigm that learns to localize UAVs using only satellite reference images, removing the need to collect paired training data. Our implementation of this paradigm, \method, is a lightweight, edge-based model that leverages a perceptual loss and non-contrastive fine-tuning to create a robust embedding space that bridges the cross-view domain gap.

Our experiments demonstrate the effectiveness of this approach. On our new benchmark, \method~achieves a high degree of precision, particularly at strict localization thresholds, while being an order of magnitude more computationally efficient than fully-supervised competitors. To facilitate this research, we also introduced the \dataset~dataset, a new and challenging benchmark featuring real-world UAV images captured at high altitudes up to 1600m. Through this work, we have shown that it is possible to achieve highly competitive localization performance in a more practical, reference-only training setting.

%%%%%%%%% REFERENCES
{\small
\bibliographystyle{ieee_fullname}
\bibliography{egbib}
}

\clearpage
\setcounter{page}{1}
\maketitlesupplementary

% \section{Rationale}
% \label{sec:rationale}
% % 
% Having the supplementary compiled together with the main paper means that:
% % 
% \begin{itemize}
% \item The supplementary can back-reference sections of the main paper, for example, we can refer to \cref{sec:intro};
% \item The main paper can forward reference sub-sections within the supplementary explicitly (e.g. referring to a particular experiment); 
% \item When submitted to arXiv, the supplementary will already included at the end of the paper.
% \end{itemize}
% % 
% To split the supplementary pages from the main paper, you can use \href{https://support.apple.com/en-ca/guide/preview/prvw11793/mac#:~:text=Delete%20a%20page%20from%20a,or%20choose%20Edit%20%3E%20Delete).}{Preview (on macOS)}, \href{https://www.adobe.com/acrobat/how-to/delete-pages-from-pdf.html#:~:text=Choose%20%E2%80%9CTools%E2%80%9D%20%3E%20%E2%80%9COrganize,or%20pages%20from%20the%20file.}{Adobe Acrobat} (on all OSs), as well as \href{https://superuser.com/questions/517986/is-it-possible-to-delete-some-pages-of-a-pdf-document}{command line tools}.

\section{\dataset \space dataset details}
\label{sec:dataset_details}
As described in \Cref{subsec: uav_images} of the main paper, our dataset \dataset \space is composed of real-world UAV images captured during two actual flights conducted by an ultralight aircraft, equipped with a drone's camera and Inertial Measurement Unit (IMU), following typical drone flight trajectories.

\subsection{First flight}

\begin{figure}[b]
    \centering
    \includegraphics[width=0.9\linewidth]{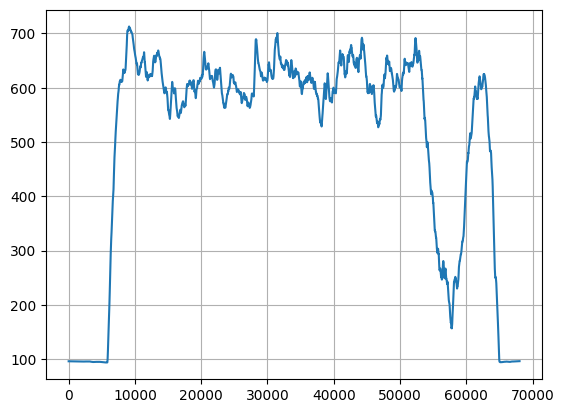}
    \caption{Altitude of drone throughout Flight 1. Altitude (m) is shown on the y-axis, and the x-axis represents each timestamp throughout the flight.}
    \label{fig: flight10_alt}
\end{figure}

The first flight, composed of 30,909 images, is the one we used to test our \method \space method.
As shown on \Cref{fig: flight10_alt}, the drone had a fairly stable altitude throughout the flight, with the exception of a couple of zones. Those induced errors for \method, as shown in \Cref{fig: vol10_traj_altitude} of the main paper.

\begin{figure}
    \centering
    \includegraphics[width=1\linewidth]{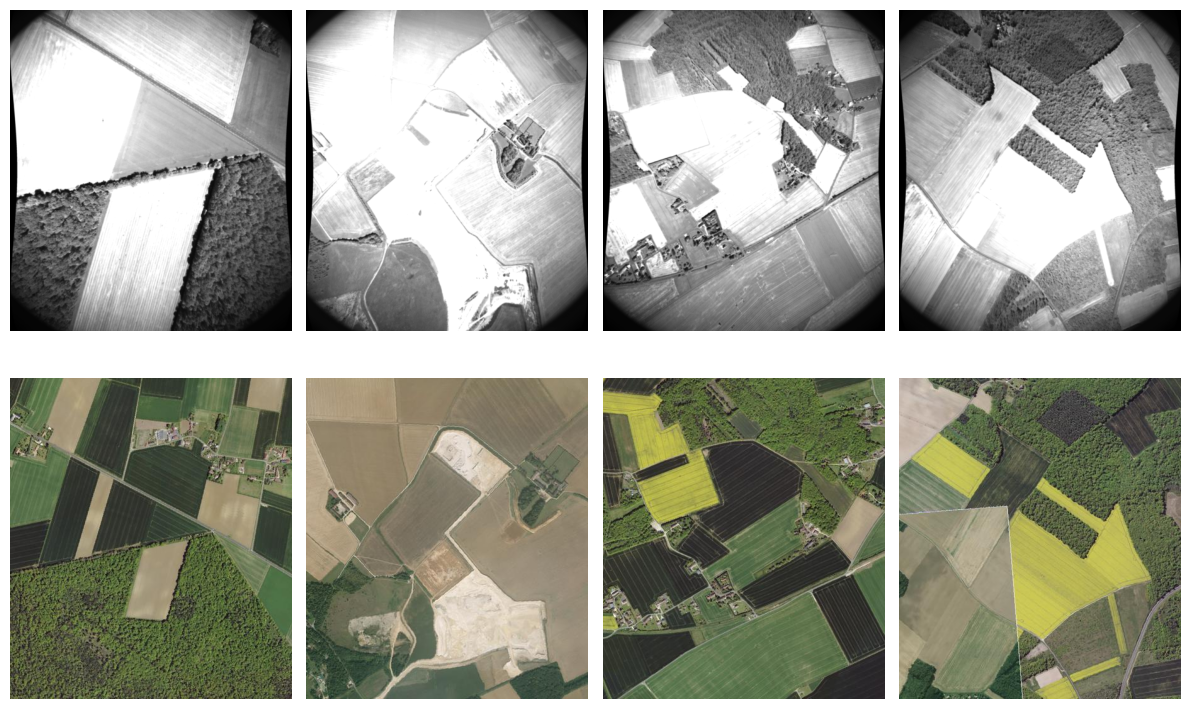}
    \caption{First set of examples of randomly picked UAV images (upper row) from Flight 1 and their geographically closest reference images (lower row).}
    \label{fig: flight10_examples}
\end{figure}

\begin{figure}
    \centering
    \includegraphics[width=1\linewidth]{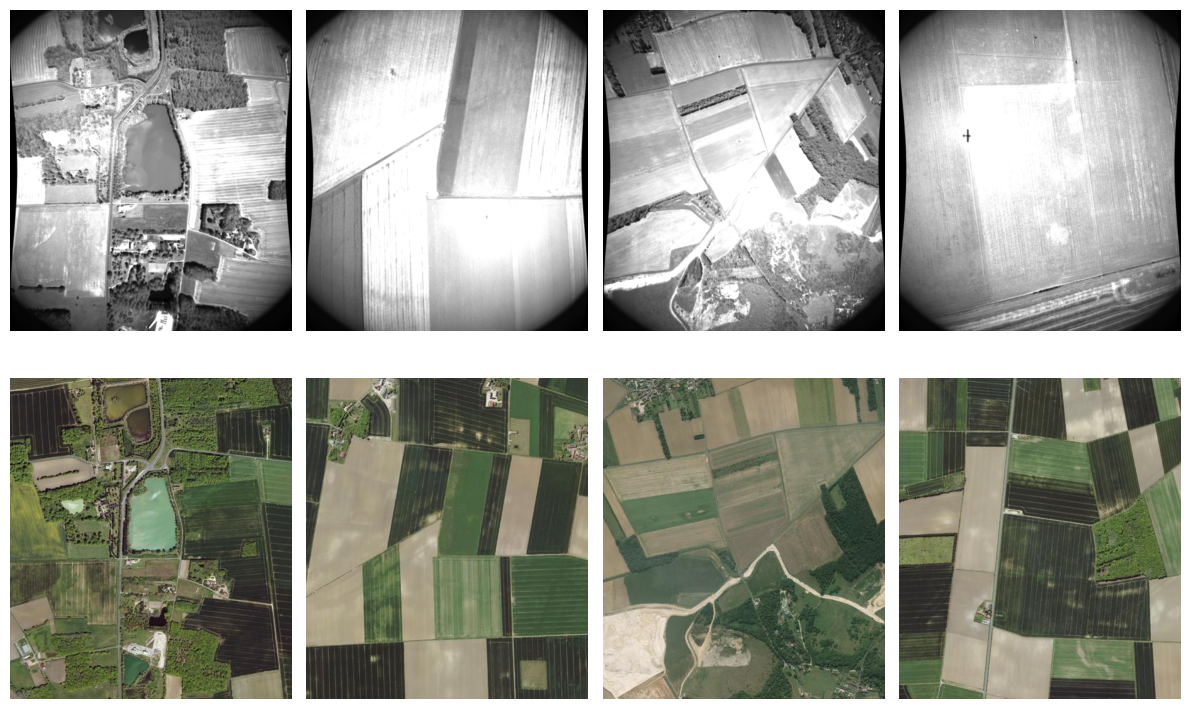}
    \caption{Second set of examples of randomly picked UAV images (upper row) from Flight 1 and their geographically closest reference images (lower row).}
    \label{fig: flight10_examples2}
\end{figure}

Figures \ref{fig: flight10_examples} and \ref{fig: flight10_examples2} show interesting examples of UAV and reference images, typical of this flight. For instance, the far left column on \Cref{fig: flight10_examples} shows the difference in viewpoints between UAV and reference images when the drone's altitude starts to drop significantly. So does the far right column of \Cref{fig: flight10_examples2}. The third column on \Cref{fig: flight10_examples2} shows the small shift in viewpoint that we can observe when the pitch or roll of the drone is not null. 

\begin{table}[t]
\centering
\caption{Coverage percentage by land cover type for Flight 1. Columns 3 and 4 also show the highest and lowest coverage percentage per image per class.}
\label{tab:land_cover_flight1}
\resizebox{0.45\textwidth}{!}{%
\begin{tabular}{l r r r}
\toprule
\textbf{Land Cover Type} & \textbf{Coverage (\%)} & \textbf{Highest (\%)} & \textbf{Lowest (\%)} \\
\midrule
Noise           & 0.311       & 38.373 & 0 \\
Low vegetation  & 0.236       & 5.016 & 0 \\
High vegetation & 2.953       & 47.391 & 0 \\
Forest          & 17.023      & 83.074 & 0 \\
Field           & 76.483      & 99.204 & 6.171 \\
Stone           & 0.0001      & 0.669 & 0 \\
River / lake    & 0.487       & 17.376 & 0 \\
Road            & 0.587       & 6.171 & 0 \\
Building        & 0.155       & 4.539 & 0 \\
Industrial zone & 0.00003     & 0.0754 & 0 \\
Railroad        & 0.0008      & 0.336 & 0 \\
Field path      & 1.760       & 7.746 & 0.0170 \\
\bottomrule
\end{tabular}%
}
\end{table}

\subsection{Second flight}

\begin{figure}
    \centering
    \includegraphics[width=1\linewidth]{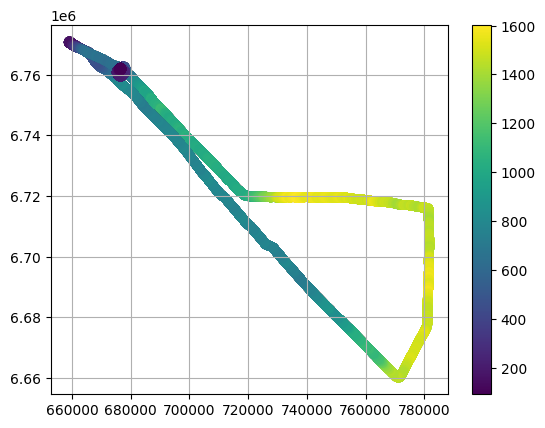}
    \caption{Illustration of the altitude of the drone throughout the second whole trajectory of flight 2, in meters}
    \label{fig: flight09_alt_traj}
\end{figure}

\begin{figure}
    \centering
    \includegraphics[width=0.9\linewidth]{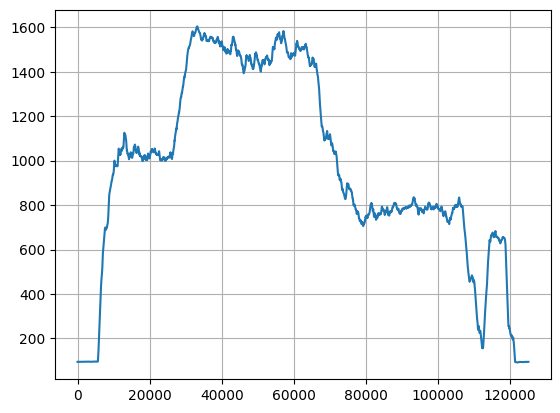}
    \caption{Altitude of drone throughout flight 2. Altitude (m) is shown on the y-axis, and the x-axis represents each timestamp throughout the flight.}
    \label{fig: flight09_alt}
\end{figure}

The second flight recorded 59,438 images. The landscapes recorded during this flight are similar to those recorded during the first flight. This flight has a less complex trajectory than Flight 1, \ie the trajectory of Flight 2 is mainly composed of straight lines rather than sharp turns like in Flight 1. However, as shown on \Cref{fig: flight09_alt_traj}, the images are recorded at higher altitudes during Flight 2, and with more altitude changes. Indeed, Flight 1 has an average altitude of 571m with a standard deviation of 116m, while Flight 2 has an average altitude of 1007m with a standard deviation of 402m. \Cref{fig: flight09_alt} shows the evolution of the drone's altitude throughout the flight. Like for the first flight, the altitude is fairly stable, but has some sudden up or down changes.

\begin{figure}
    \centering
    \includegraphics[width=1\linewidth]{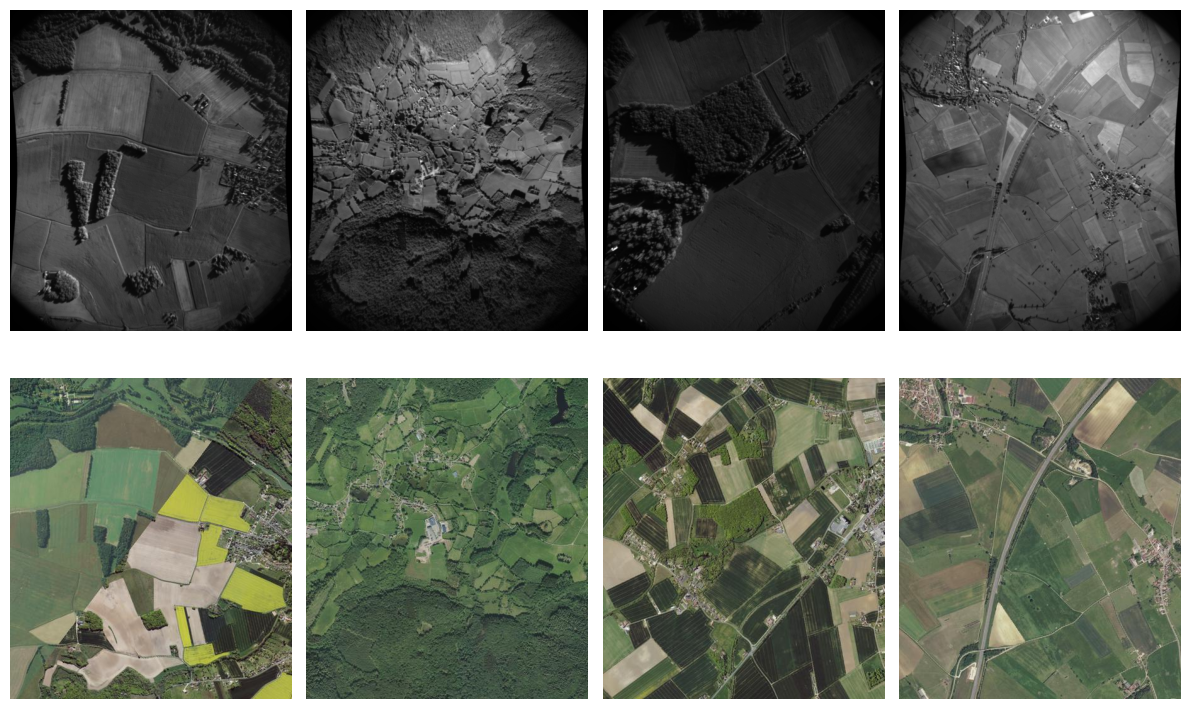}
    \caption{First set of examples of randomly picked UAV images (upper row) from Flight 2 and their geographically closest reference images (lower row).}
    \label{fig: flight09_examples}
\end{figure}

\begin{figure}
    \centering
    \includegraphics[width=1\linewidth]{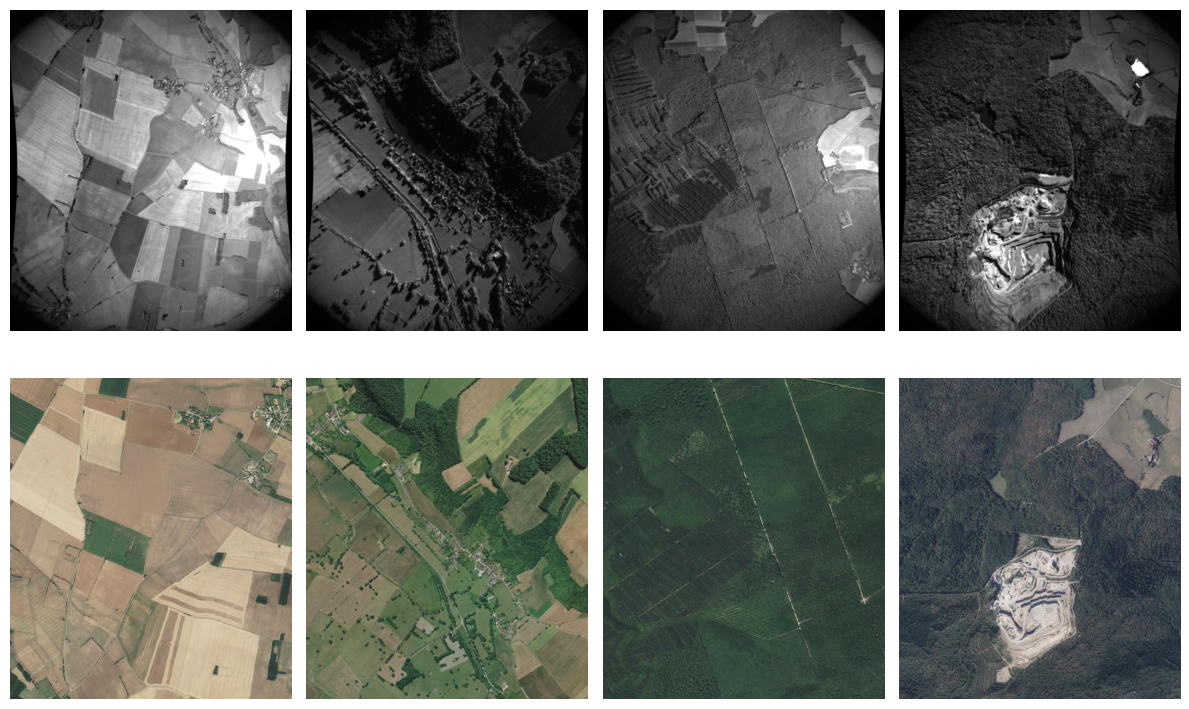}
    \caption{Second set of examples of randomly picked UAV images (upper row) from Flight 2 and their geographically closest reference images (lower row).}
    \label{fig: flight09_examples2}
\end{figure}

\begin{figure}
    \centering
    \includegraphics[width=1\linewidth]{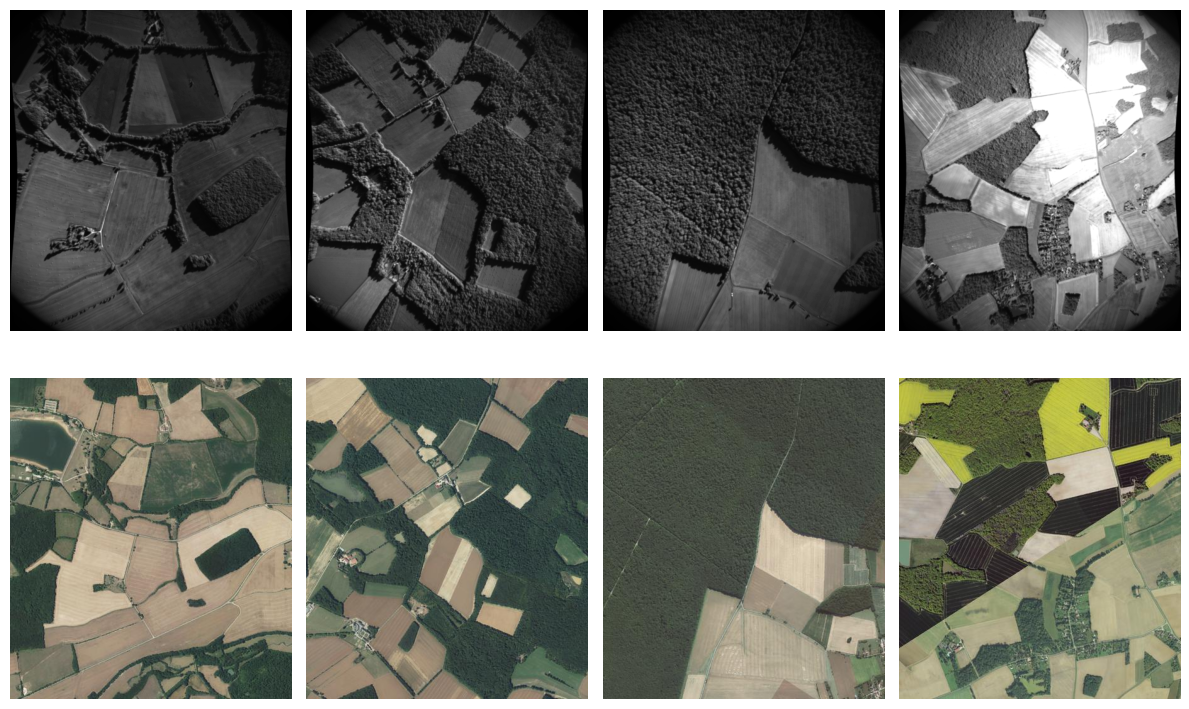}
    \caption{Third set of examples of randomly picked UAV images (upper row) from Flight 2 and their geographically closest references images (lower row).}
    \label{fig: flight09_examples3}
\end{figure}

Figures \ref{fig: flight09_examples}, \ref{fig: flight09_examples2} and \ref{fig: flight09_examples3} show random examples of UAV images from Flight 2, as well as the closest reference images. Those examples show that the hypothesis of null pitch and roll holds better for this flight, as the trajectory has more straight lines and fewer turns. However, the larger altitude range creates bigger differences in viewpoints between UAV and reference images. 

\subsection{Differences with existing UAV datasets}
\begin{figure}
    \centering
    \includegraphics[width=1\linewidth]{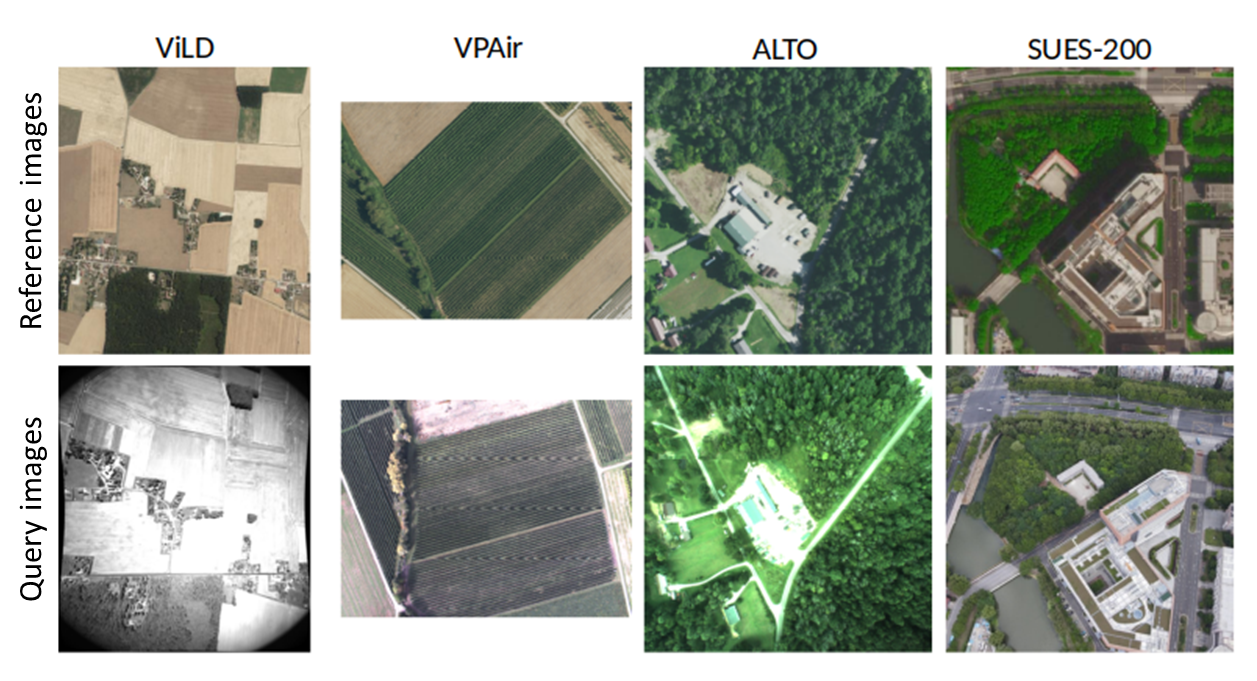}
    \caption{Comparison of a randomly selected query-reference pair from different datasets.}
    \label{fig:vpr_datasets_comparison}
\end{figure}

\begin{figure}
    \centering
    \includegraphics[width=1\linewidth]{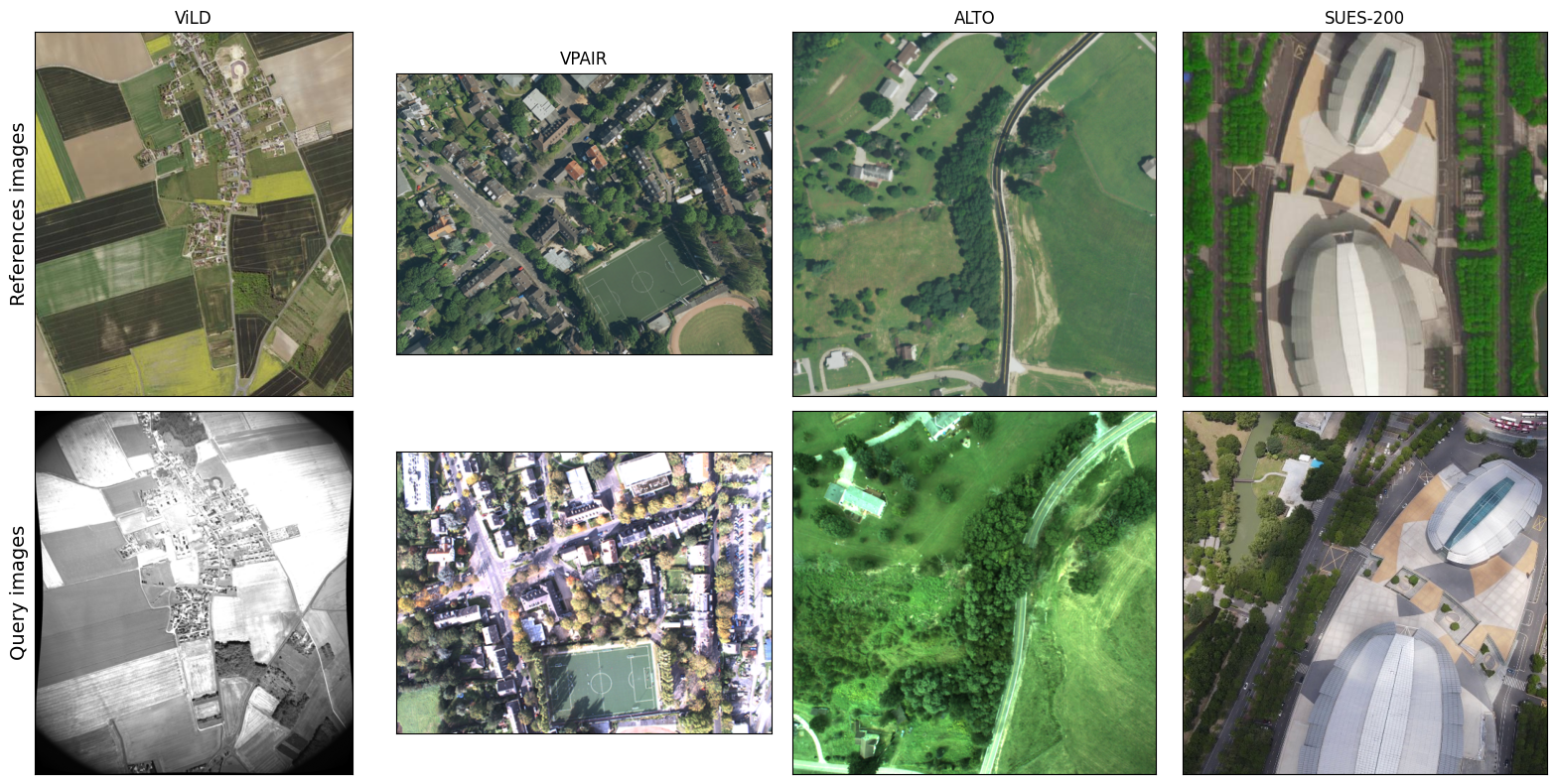}
    \caption{Comparison of a randomly selected query-reference pair from different datasets.}
    \label{fig:vpr_datasets_comparison_1}
\end{figure}

We talk, in \Cref{subsec:diff_datasets} of the main paper, about the differences between \dataset~and other UAV datasets in the literature. We show here, with Figures \ref{fig:vpr_datasets_comparison} and \ref{fig:vpr_datasets_comparison_1}, some examples of reference-query image pairs from different datasets to underline the said differences. 

\section{Implementation details}
\label{sec:supp__implem_details}
As explained in \Cref{sec: experiments}, the autoencoder architecture is based on~\cite{hou2017deep} and~\cite{dipiazza2024leveragingedgedetectionneural}. The encoder consists of four strided convolutional layers with batch normalization and \textit{LeakyReLU} activation, producing a 1024-dimensional latent space. Only the encoder is deployed on the UAV, and it contains 33 million parameters.
The decoder mirrors the encoder’s structure, with a \textit{Sigmoid} activation on its final layer to accommodate the binary contour input images. In \cref{eq: perceptual}, $\beta$ is set to 1. The global projection head consists of 2 linear layers with batch normalization and \textit{ReLU}, followed by a third linear layer, all of size 4096. The local projection head has the same structure, but each layer is sized at 560. Following~\cite{bardes2022vicreglselfsupervisedlearninglocal}, $\alpha$ in \cref{eq: vicregl} is 0.75, and $\gamma$ is set to 20 for the top-$\gamma$ pairs retained in \cref{eq: loc_based} and (\ref{eq: feature_based}). Models were trained on reference images  for 100 epochs using the AdamW optimizer and then tested on the UAV images.

The vignetting augmentation was applied by multiplying the input image with a centered Gaussian filter matrix, matching the image size, with a standard deviation of 70. The translation offset was selected uniformly within [-$x$, $x$], where $x$ is the pixel distance equivalent to 10 meters in real life. Rotations were randomly chosen between -30° and 30°, and cropping was centered on a random point, with scale set between 70\% and 100\%. Noise and blur were added using a Gaussian distribution and a kernel size of 5, while brightness and contrast factors varied uniformly between 0 and 2. All augmentations were applied to grayscale images prior to Canny edge extraction.

\section{Experiments and analyses}
\label{sec:supp_experiments}

\subsection{Altitude drops}
\label{subsec:supp__alt_drops}

As shown on \Cref{fig: vol10_traj_altitude} and discussed in \Cref{subsec: implementation_details} of the main paper, \method \space can handle moderate altitude variations when localizing query images, but large altitude changes make the method struggle to correctly localize UAV images. To assess this phenomenon, we calculated localization scores for images captured above 300m, 400m, 450m, and 500m.
\begin{figure}
    \centering
    \includegraphics[width=1\linewidth]{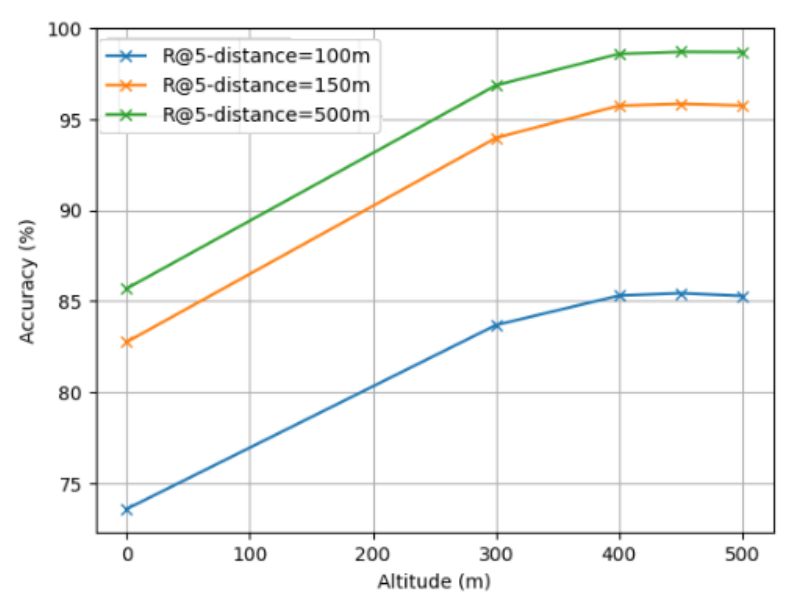}
    \caption{Localization performance when removing images below a given altitude.}
    \label{fig: scores_per_altitude_drop}
\end{figure}
As shown in \Cref{fig: scores_per_altitude_drop}, while \method \space remains robust to drops of 100-200m, performance declines when altitude drops exceed 200m. Excluding images below 300m or 500m minimally impacts model performance.

\subsection{Localization performance on a subset of queries}
\label{subsec:supp__loc_perf_subset}

To better compare the localization performance of SuperPoint~+~LightGlue to the performance of the methods reported in \Cref{tab:recall_comparison}, we reproduce the same experiment that we did for SuperPoint~+~LightGlue. Indeed, as we explained in \Cref{tab:recall_comparison}, localizing one query image against the whole reference database would take us a very long time with SuperPoint~+~LightGlue. As we still wanted to be able to compare the results of this method against other vector-based methods, we decided to sample 500 query images from the test set. To localize each of the 500 query images, we gathered all the reference images that were less than 1km away from the query image, and we localized the query image only against those close reference images. Results are shown on \Cref{tab:recall_comparison__subset}.

\method \space consistently outperforms SuperPoint~+~LightGlue up to 250m. At the 500m mark, the performance levels converge, which is expected given that the search is limited to a 1km radius around each query image. Interestingly, SuperPoint~+~LightGlue achieves slightly better results for R@5 and R@10 at 500m, though at the cost of a much longer computational time. This advantage may come from its strong robustness to extreme viewpoint variations, such as large heading and altitude differences, among others. While this may help in retrieving database images that resemble the query image ---\ie bumping up the localization performance at 500m---, it could also make it harder for the method to precisely identify the closest match, as it remains invariant to these transformations. 

\begin{table*}[ht]
\centering
\resizebox{0.99\textwidth}{!}{%
\begin{tabular}{@{}l c ccc ccc ccc ccc c c@{}}
\toprule
& \multirow{2}{*}{\shortstack{\textbf{Satellite-Only} \\ \textbf{Training}}}  & \multicolumn{3}{c}{\textbf{100m}} & \multicolumn{3}{c}{\textbf{150m}} & \multicolumn{3}{c}{\textbf{250m}} & \multicolumn{3}{c}{\textbf{500m}} & \multirow{2}{*}{\shortstack{\textbf{Descriptors} \\ \textbf{Dimension}}} & \multirow{2}{*}{\shortstack{\textbf{GFLOPs} \\ \textbf{per query}}}\\
\cmidrule(lr){3-5} \cmidrule(lr){6-8} \cmidrule(lr){9-11} \cmidrule(lr){12-14}
Method & & R@1 & R@5 & R@10 & R@1 & R@5 & R@10 & R@1 & R@5 & R@10 & R@1 & R@5 & R@10 &  & \\ 
\midrule
\midrule

Random & - & 0.03 & 0.16 & 0.33 & 0.07 & 0.36 & 0.73 & 0.21 & 1.02 & 2.03 & 0.86 & 4.16 & 8.06 & - & - \\

\midrule

SuperPoint~\cite{detone2018superpoint}~+~LightGlue~\cite{lightglue}* & \xmark & 11.60 & 33.80 & 48.40 & 24.80 & 54.00 & 67.40 & 49.40 & 78.20 & 86.20 & 86.20 & 95.60 & \underline{97.60} & 256 & 17,705.44 \\

\midrule

MixVPR~\cite{ali2023mixvpr} (Zero-shot) & \multirow{2}{*}{\xmark} & 15.40 & 36.60 & 52.40 & 29.40 & 55.80 & 69.60 & 59.20 & 78.00 & 86.00 & 87.60 & 91.40 & 93.60 & \multirow{2}{*}{4096} & \multirow{2}{*}{\underline{10.31}} \\
MixVPR (Finetuned) & & 45.00 & 68.80 & 78.20 & 69.40 & 83.80 & 87.40 & 82.60 & 89.20 & 91.40 & 90.40 & 94.40 & 95.00 \\ [2mm]

EigenPlaces~\cite{eigenplaces} (Zero-shot) & \multirow{2}{*}{\xmark} & 15.80 & 35.60 & 49.00 & 27.80 & 52.00 & 62.80 & 49.00 & 71.20 & 79.20 & 81.40 & 91.80 & 94.80 & \multirow{2}{*}{2048} & \multirow{2}{*}{19.71} \\
EigenPlaces (Finetuned) & & 42.40 & 72.40 & 82.20 & 63.80 & 86.00 & 89.80 & 84.00 & 92.00 & 93.80 & 90.80 & 92.60 & 95.60 \\ [2mm]

Megaloc~\cite{berton_2025_megaloc} (Zero-shot) & \xmark & 5.60 & 17.60 & 24.80 & 10.60 & 30.60 & 40.00 & 29.80 & 54.80 & 65.60 & 66.20 & 83.80 & 89.40 & 8448 & 54.93\\

\midrule

FSRA~\cite{dai2021transformer} (Zero-shot) & \multirow{2}{*}{\xmark} & 2.00 & 12.00 & 18.60 & 5.20 & 20.80 & 32.40 & 17.60 & 10.80 & 53.20 & 53.80 & 80.00 & 88.00 & \multirow{2}{*}{\textbf{512}} & \multirow{2}{*}{13.34} \\
FSRA (Finetuned) & & 38.00 & \underline{74.00} & 80.60 & 67.40 & 83.80 & \textbf{86.60} & \underline{89.20} & \underline{91.60} & \underline{92.60} & \underline{93.80} & \underline{95.60} & \textbf{97.80} \\ [2mm]

DAC~\cite{xia2024enhancing} (Zero-shot) & \multirow{2}{*}{\xmark} & 2.80 & 11.40 & 15.60 & 5.40 & 17.00 & 24.80 & 15.20 & 33.20 & 45.20 & 45.60 & 68.20 & 78.60 & \multirow{2}{*}{\underline{1024}} & \multirow{2}{*}{20.55} \\
DAC (Finetuned) & & \textbf{52.60} & \textbf{74.80} & \textbf{82.40} & \textbf{79.20} & \textbf{85.20} & \textbf{86.60} & \textbf{91.80} & \textbf{93.00} & \textbf{93.40} & \textbf{94.2} & \textbf{96.20} & \textbf{97.80} &  \\

\midrule

Di~Piazza~\etal~\cite{dipiazza2024leveragingedgedetectionneural} (Finetuned) & \checkmark & 34.00 & 48.20 & 53.00 & 42.80 & 56.20 & 60.40 & 46.40 & 61.20 & 66.40 & 55.80 & 75.00 & 81.80 & \underline{1024} & \textbf{1.42} \\ 

\midrule

\textbf{\method \space (Ours)} & \checkmark & \underline{51.20} & \textbf{74.80} & \underline{81.40} & \underline{74.60} & \underline{84.80} & \underline{86.00} & 85.40 & 88.40 & 88.60 & 91.80 & 93.20 & {97.20} & \underline{1024} & \textbf{1.42} \\

\bottomrule
\end{tabular}%
}
\caption{Comparison of Recall@K performance for various methods at different localization thresholds (100m, 150m, 250m, and 500m). The performance reported in this table is based on a subset of 500 query images, rather than the entire set of query images. Each query is only matched against reference images within a 1km radius. This setup ensures a fair comparison between all methods and SuperPoint~+~LightGlue. As noted in \Cref{tab:recall_comparison}, localizing the full query set against all reference images would be too time-consuming for SuperPoint and LightGlue, which is why we limited the search space.}
\label{tab:recall_comparison__subset}
\end{table*}

\subsection{Robustness to image perturbations}
To assess whether the learned model overly relies on the synthetic spatial warping augmentations used during training, we evaluate its robustness to different perturbations applied to the reference images at inference time.
Specifically, we consider three types of perturbations:
\begin{itemize}
    \item Rotation, where the image is randomly rotated within a range of 0° to 60° to simulate the variations in UAV heading throughout a flight.
    \item Center crop and resize, where the image is cropped and resized to simulate variations in UAV altitude, with crop factors ranging from 100\% (no crop) to 50\% of the original image.
    \item Color jittering, where random changes in brightness, contrast, saturation, and hue are applied with parameters ranging from (0,0,0,0) to (0.4,0.4,0.4,0.1).
\end{itemize}

For each perturbation type and severity level, we perform five runs, each time applying the perturbations at random to the reference images. This stochastic evaluation ensures that the reported results are not dependent on a single perturbation configuration and provides a robust estimate of the model’s performance under challenging conditions.

\begin{figure*}
    \centering
    
        \setlength{\tabcolsep}{2pt}
    
    \begin{tabular}{cccc} 
        % \toprule
        % \multicolumn{4}{c}{\textbf{Recall@1 at 100m}} \\ 
        % \midrule

        \includegraphics[width=0.3\textwidth]{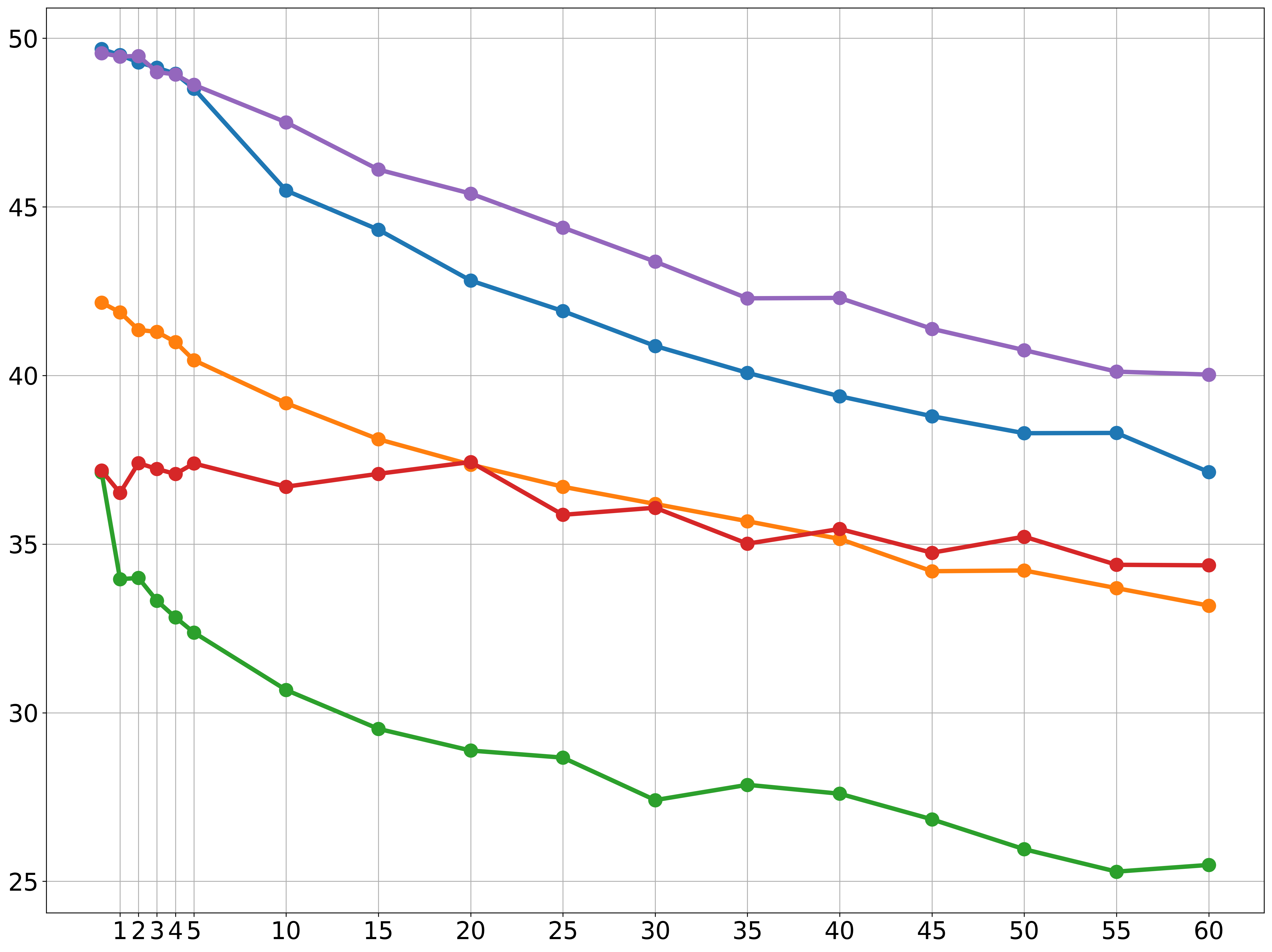} &
        \includegraphics[width=0.3\textwidth]{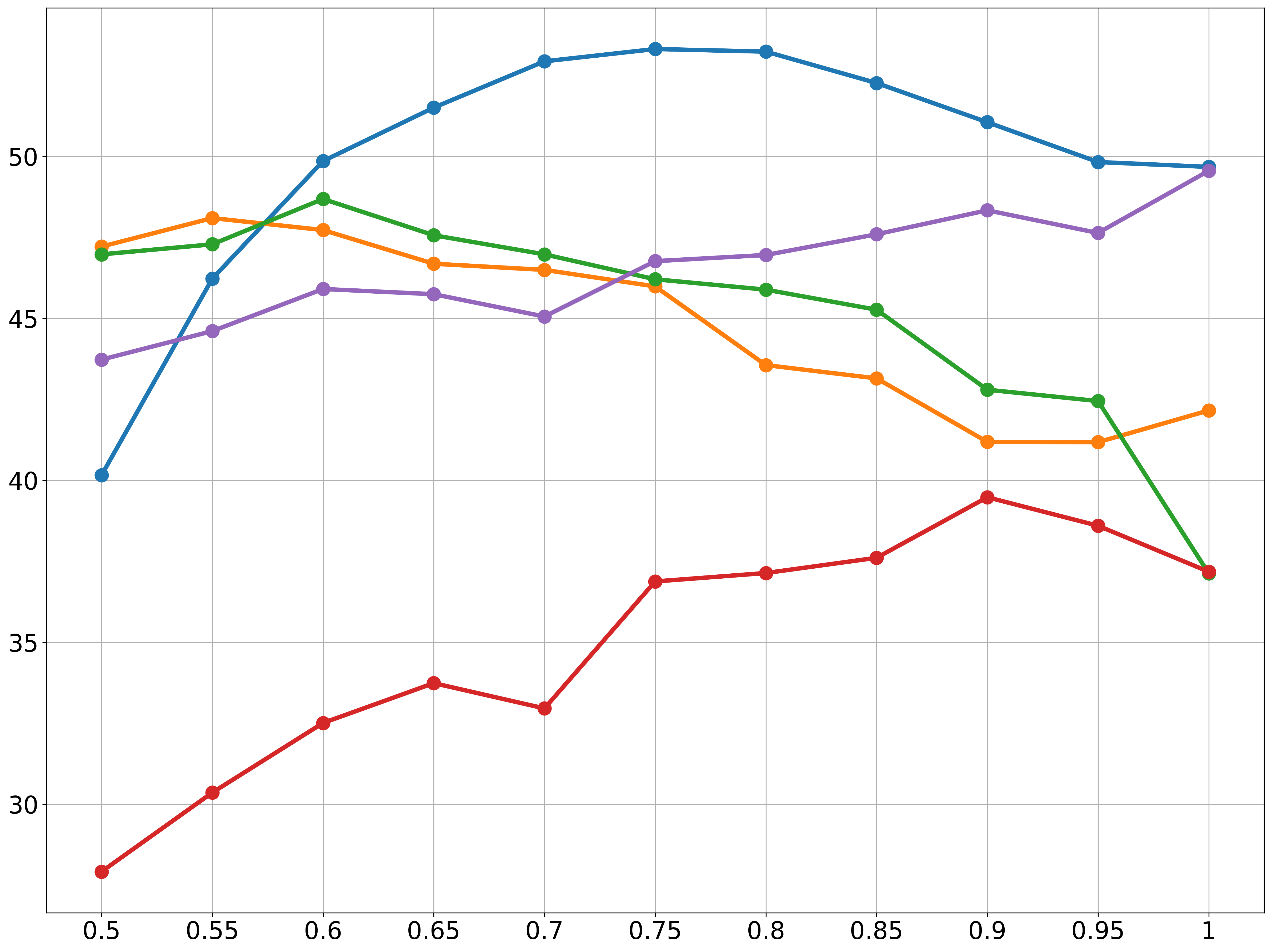} &
        \includegraphics[width=0.3\textwidth]{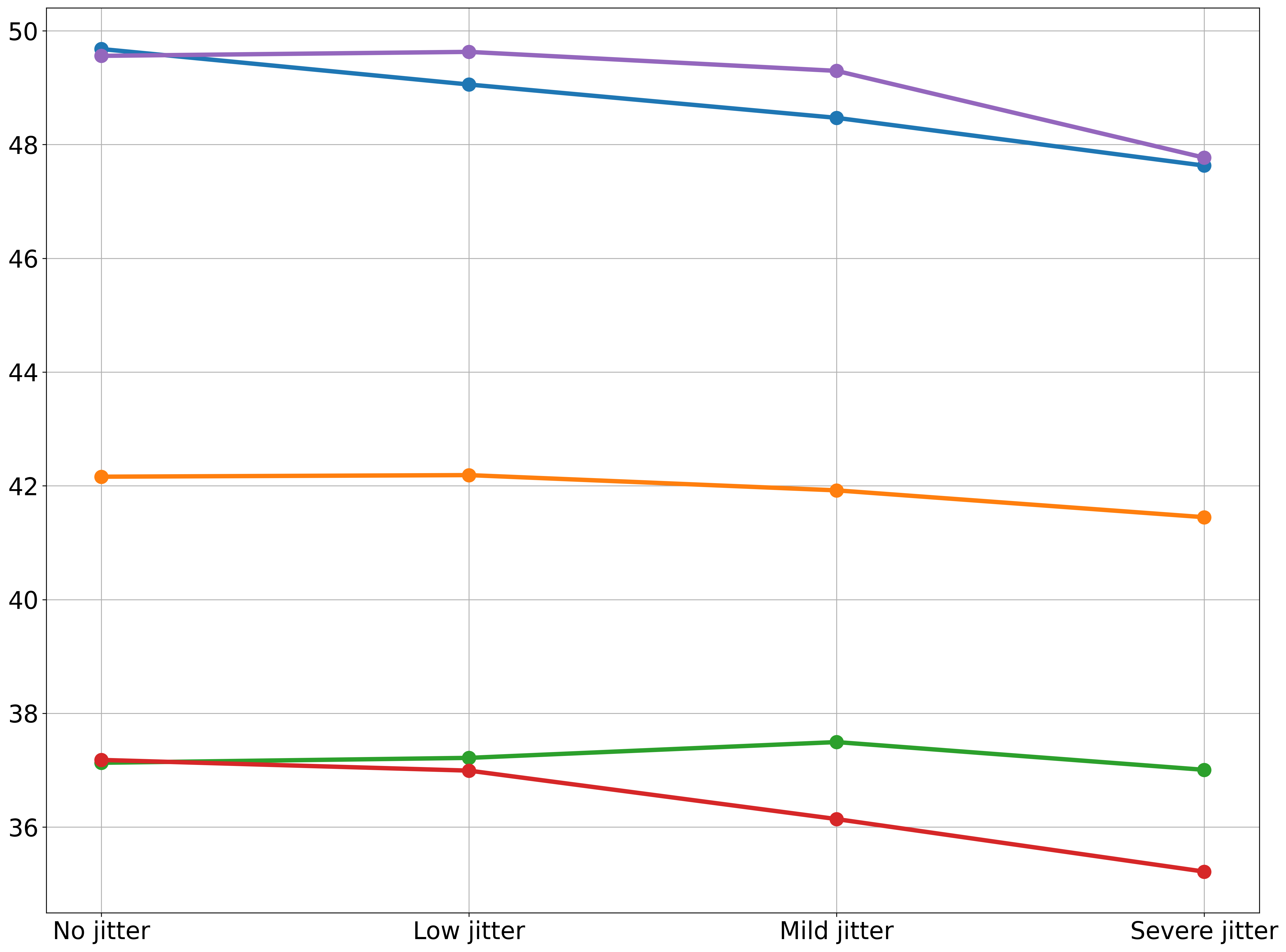} &
        \includegraphics[width=0.1\textwidth]{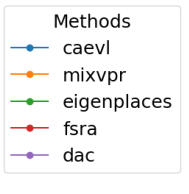} 
        \\

        Rotation & Altitude drop & Color Jitter \\

        % \bottomrule
    \end{tabular}
    \caption{Evolution of Recall@1 at 100 m for \method~and four SOTA methods under perturbations applied to the reference images.}
    \label{fig: perturbations}
\end{figure*}

\Cref{fig: perturbations} shows that \method's ability to adapt to perturbed images is on par with the other methods we compare it to. 

Interestingly, for the perturbation simulating altitude drops, the performance of \method peaks at a 75\% center crop before declining at smaller crops. We attribute this to two complementary effects. First, moderate cropping removes peripheral regions that often contain noisy or irrelevant edges, yielding cleaner Canny maps and forcing the model to focus on stable, central structures such as roads and buildings. However, when the crop becomes too aggressive (e.g., 50\%), important structural context is lost, as roads and intersections are truncated, reducing the distinctiveness of the representation. This explains the bell-shaped performance curve observed in our experiments.

\subsection{Edge detector impact}
While our method primarily uses the Canny operator for edge detection, we conducted an ablation study to evaluate other detectors, including the classical Sobel and LSD~\cite{von2012lsd}, as well as two deep-learning-based methods: HED~\cite{xie2015holistically} and DexiNed~\cite{soria2023dense}.

\begin{table}[t]
\centering
\resizebox{\linewidth}{!}{
\begin{tabular}{lccc}
\toprule
\textbf{Edge Detector} & \textbf{Recall@1 @100m} & \textbf{Recall@1 @150m} & \textbf{Time (ms)}\\ 
\midrule
Sobel     & 14.26 & 25.07 & \textbf{1.25} \\ 
LSD       & 44.58 & 69.83 & 12.5 \\ 
HED       & 48.47 & 71.45 & 26.3 \\ 
DexiNed   & \textbf{52.27} & \textbf{74.44} & 35.9  \\
Canny (ours) & \underline{49.68} & \underline{73.93} & \underline{1.49} \\ 
\bottomrule
\end{tabular}
}
\caption{Comparison of different edge detectors in our pipeline. 
Recall@1 is reported at 100m and 150m thresholds. 
Canny provides the best balance between performance and efficiency.}
\label{tab:edge_detectors}
\end{table}

\Cref{tab:edge_detectors} reports the results. Deep edge detectors like DexiNed slightly improve retrieval accuracy but are significantly slower. Canny achieves a good balance between accuracy and computational efficiency, which justifies its use in our main experiments.
Furthermore, several prior works in UAV localization have also relied on Canny for edge detection \cite{dipiazza2024leveragingedgedetectionneural, afolabi15, si2024gan}, reinforcing its relevance as a baseline for this task, which was the first reason that led us to choose it.

\begin{figure*}
    \centering
    \includegraphics[width=1\linewidth]{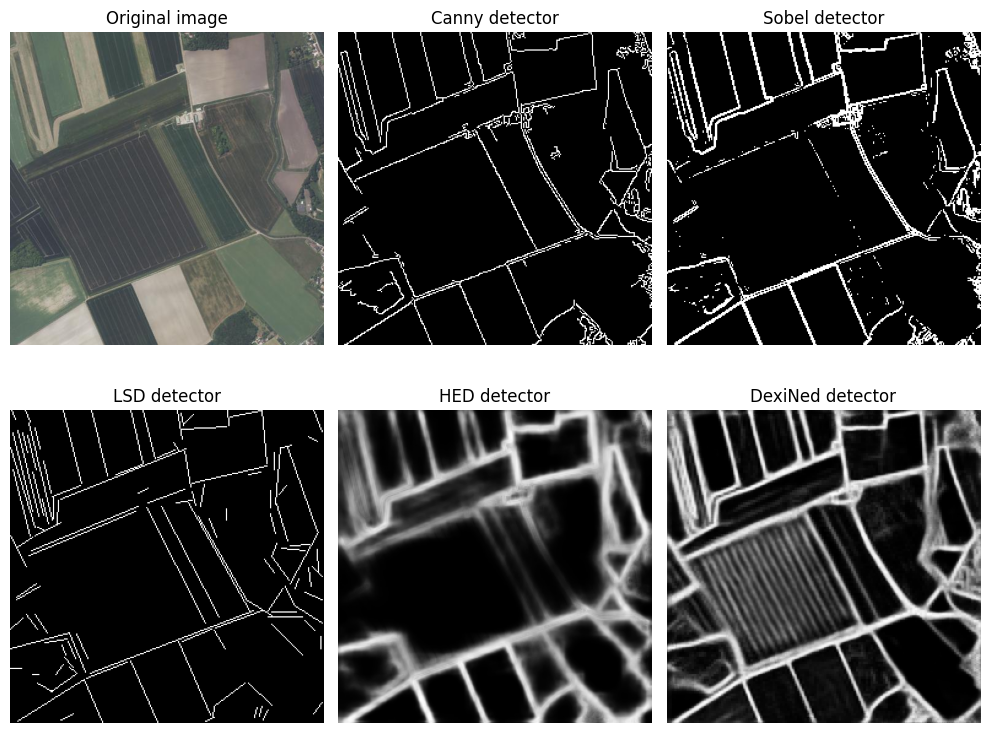}
    \caption{Randomly selected image from \dataset, processed by each edge detector considered in the ablation study.}
    \label{fig: db_vild_287525}
\end{figure*}

\begin{figure*}
    \centering
    \includegraphics[width=0.8\linewidth]{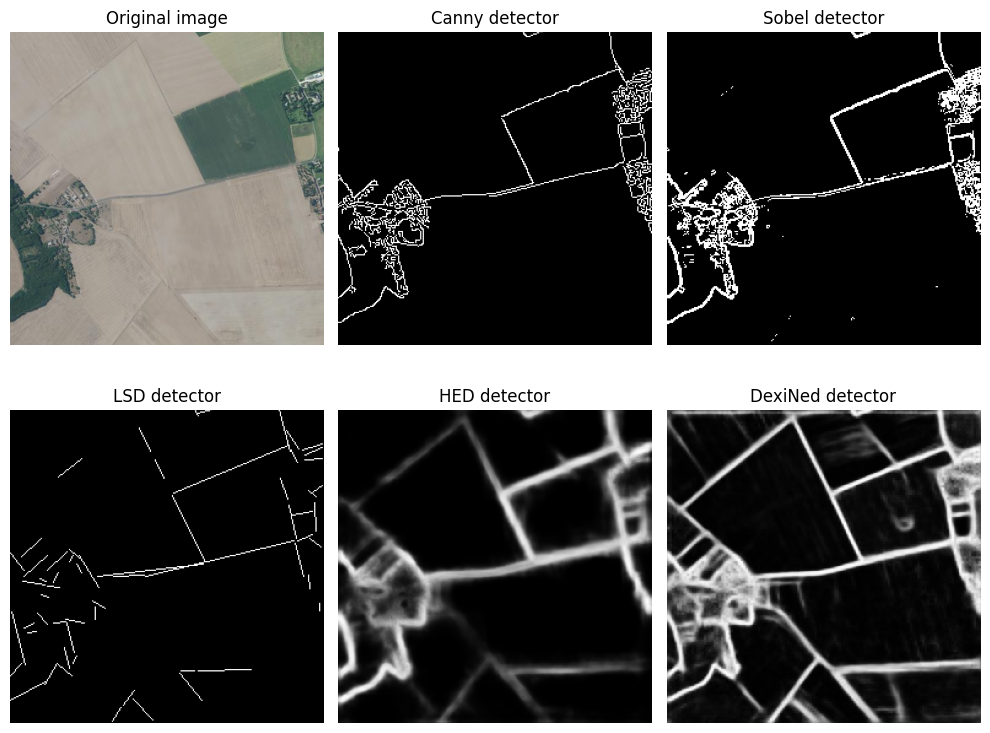}
    \caption{Randomly selected image from \dataset, processed by each edge detector considered in the ablation study.}
    \label{fig: db_vild_74339}
\end{figure*}

\begin{figure*}
    \centering
    \includegraphics[width=0.8\linewidth]{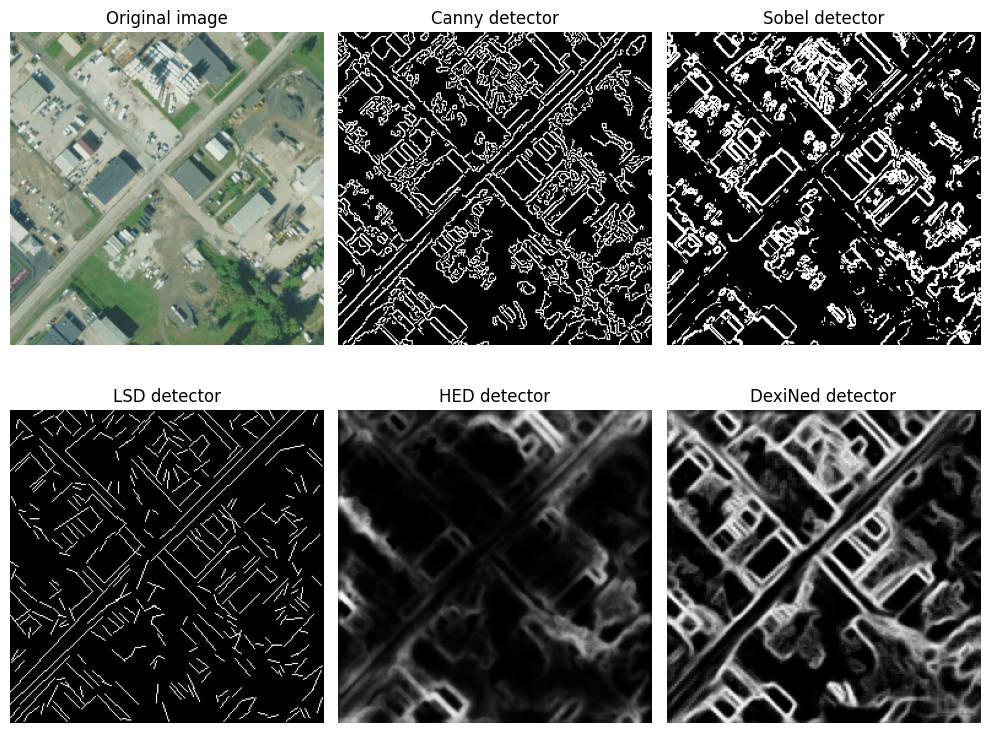}
    \caption{Randomly selected image from the Alto dataset, processed by each edge detector considered in the ablation study.}
    \label{fig: db_alto_104}
\end{figure*}

\begin{figure*}
    \centering
    \includegraphics[width=0.8\linewidth]{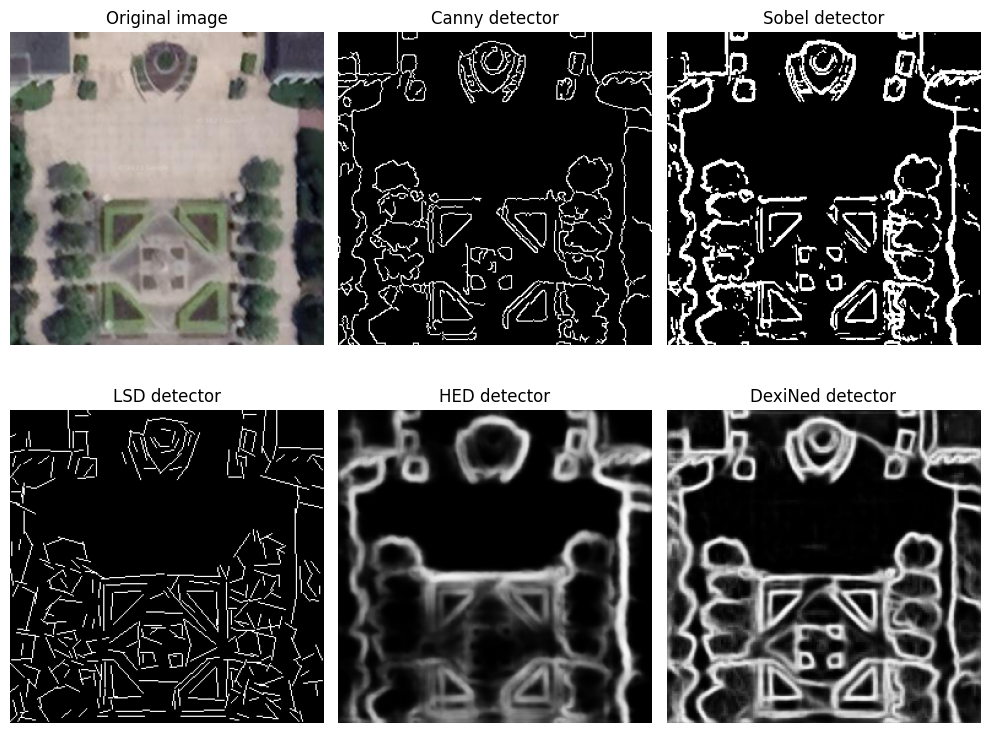}
    \caption{Randomly selected image from the DenseUAV dataset, processed by each edge detector considered in the ablation study.}
    \label{fig: db_dense_2545}
\end{figure*}

\begin{figure*}
    \centering
    \includegraphics[width=0.8\linewidth]{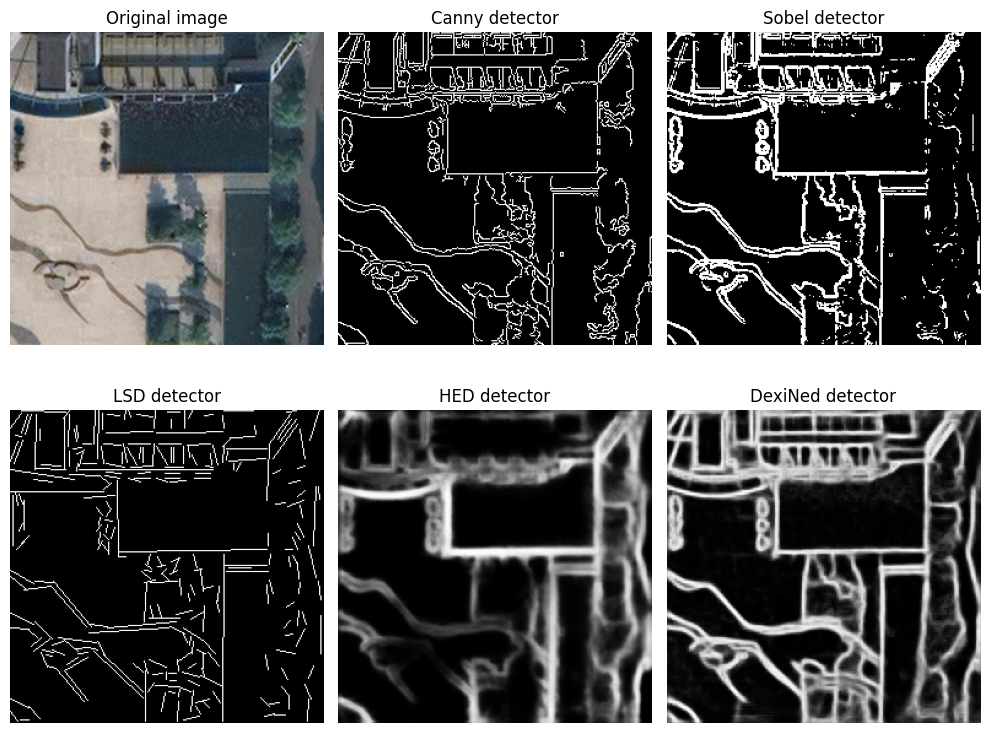}
    \caption{Randomly selected image from the DenseUAV dataset, processed by each edge detector considered in the ablation study.}
    \label{fig: db_dense_1427}
\end{figure*}

Another advantage of Canny is the balance it achieves in the amount of visual information preserved. Detectors such as LSD tend to remove too much structural detail, which can cause a loss of discriminative features. Conversely, deep-learning-based detectors like HED or DexiNed often produce dense edge maps, potentially introducing noise and spurious details that may confuse the model during training and retrieval. Canny provides a middle ground, retaining the most salient structures while discarding less informative textures, which aligns well with the goal of learning robust and generalizable image embeddings for UAV localization. 
Figures \ref{fig: db_vild_287525}, \ref{fig: db_vild_74339}, \ref{fig: db_alto_104}, \ref{fig: db_dense_2545} and \ref{fig: db_dense_1427} show a few examples of images, from \dataset~and other datasets such as DenseUAV and ALTO, that were processed by each edge detector. 

\subsection{Predictions examples}
\label{subsec: supp__pred_examples}

\begin{figure*}
    \centering
    \includegraphics[width=1\linewidth]{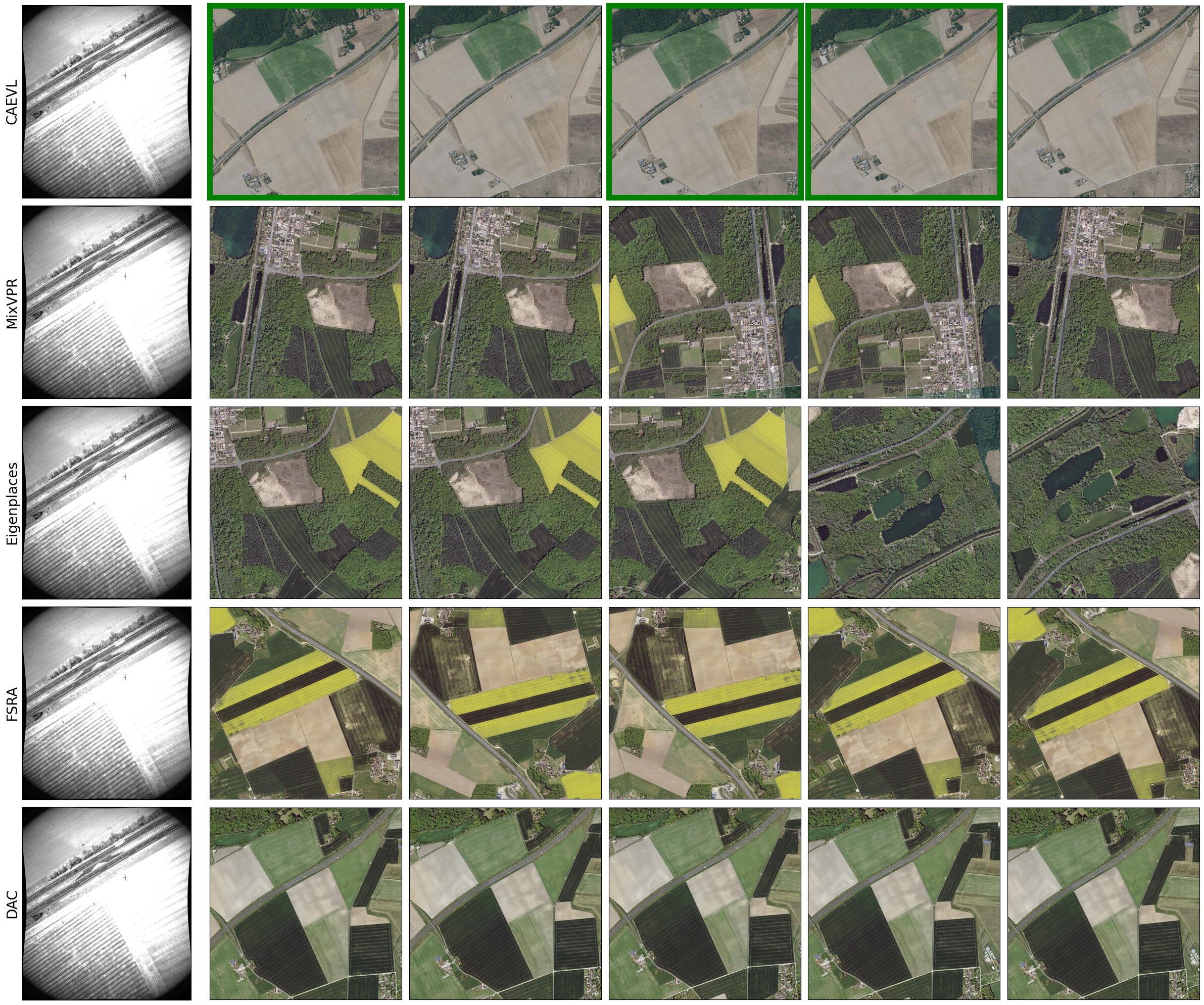}
    \caption{Visual comparison of top-5 retrieval results on a query from \dataset. For each method, the five best matches are shown to the right of a randomly picked query image. Correct predictions, defined as those within 100m of the ground truth, are highlighted with a green frame.}
    \label{fig: query_7261}
\end{figure*}

\begin{figure*}
    \centering
    \includegraphics[width=1\linewidth]{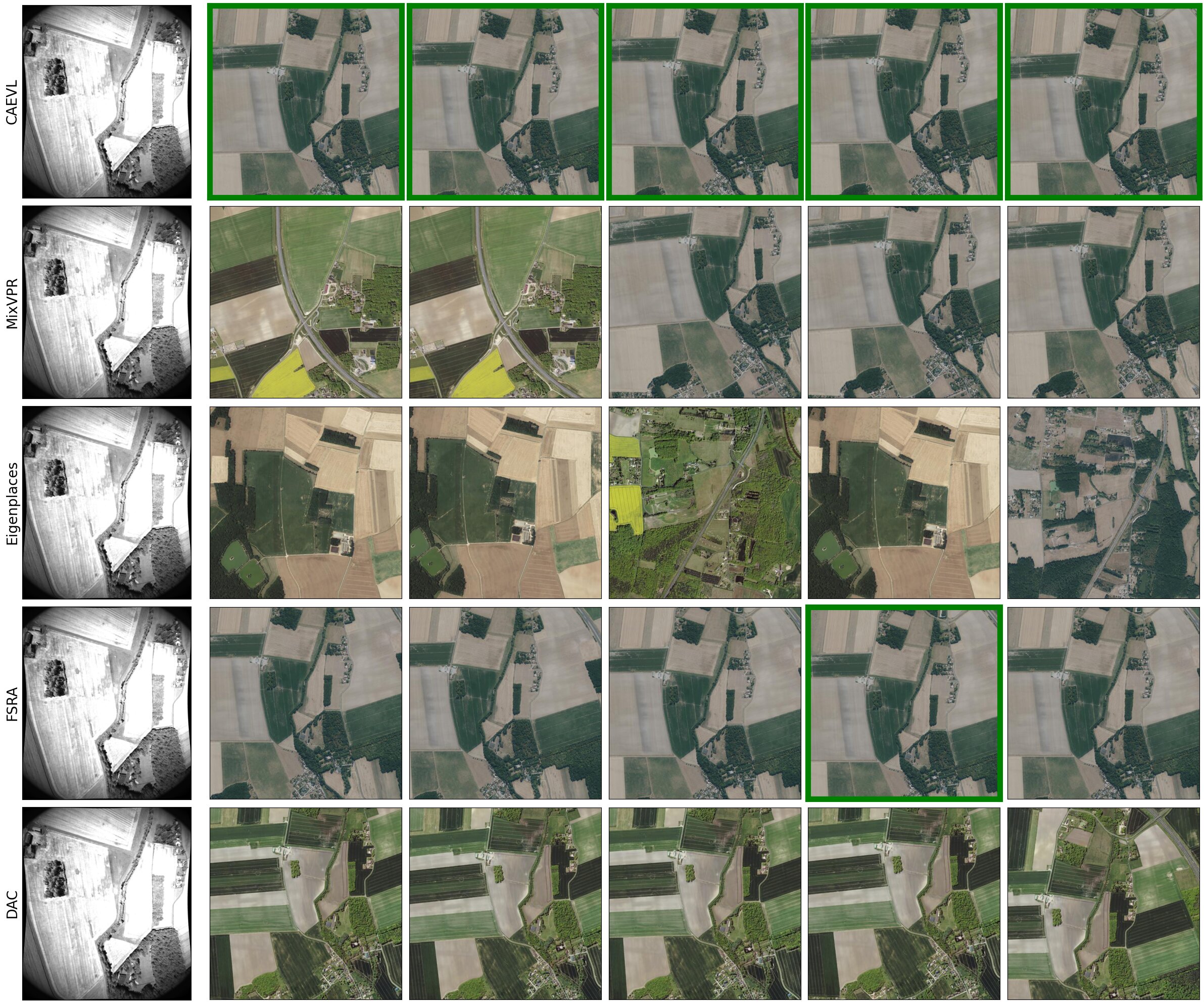}
    \caption{Visual comparison of top-5 retrieval results on a query from \dataset. For each method, the five best matches are shown to the right of a randomly picked query image. Correct predictions, defined as those within 100m of the ground truth, are highlighted with a green frame.}
    \label{fig: query_9071}
\end{figure*}

\begin{figure*}
    \centering
    \includegraphics[width=1\linewidth]{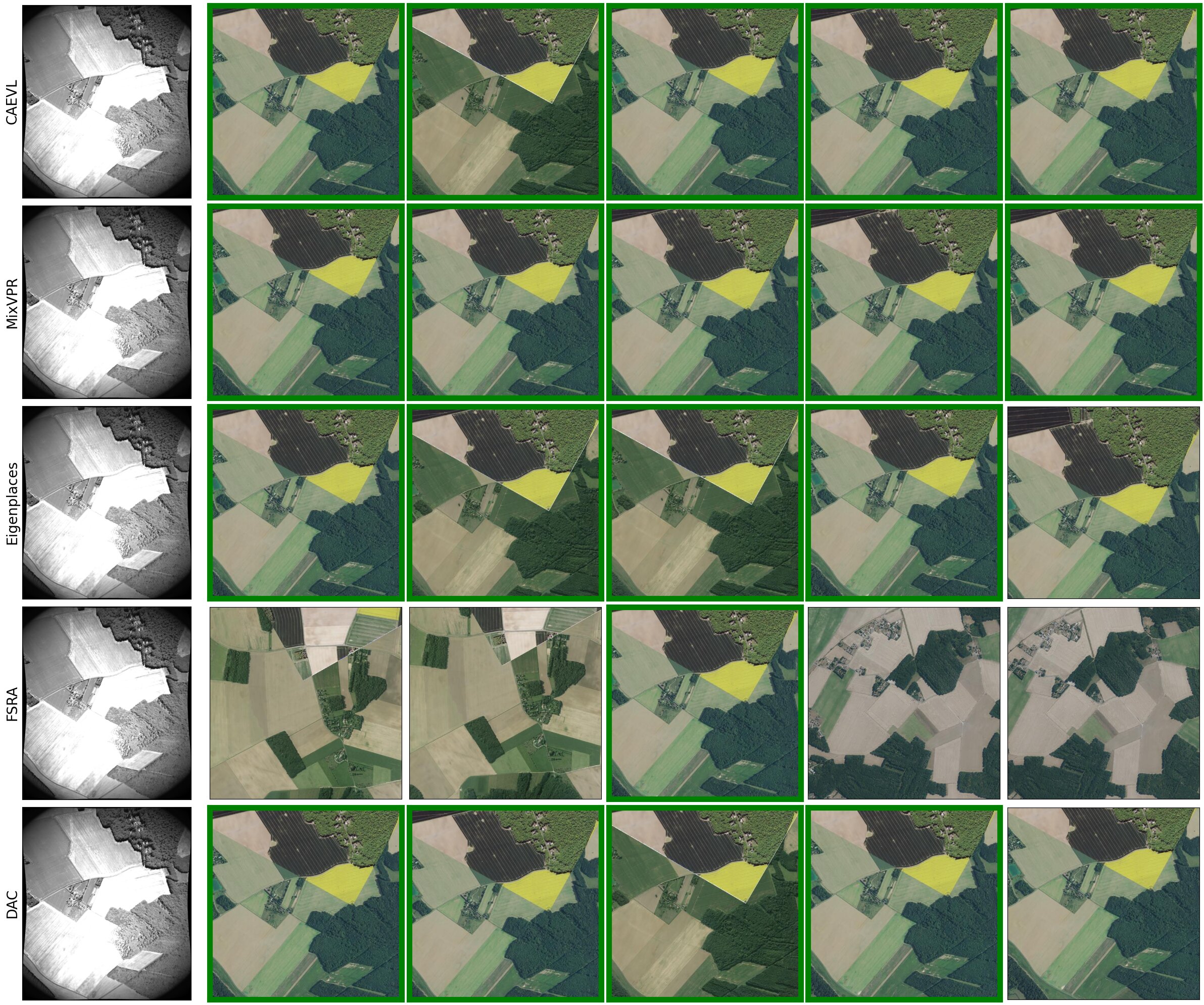}
    \caption{Visual comparison of top-5 retrieval results on a query from \dataset. For each method, the five best matches are shown to the right of a randomly picked query image. Correct predictions, defined as those within 100m of the ground truth, are highlighted with a green frame.}
    \label{fig: query_24501}
\end{figure*}

\begin{figure*}
    \centering
    \includegraphics[width=1\linewidth]{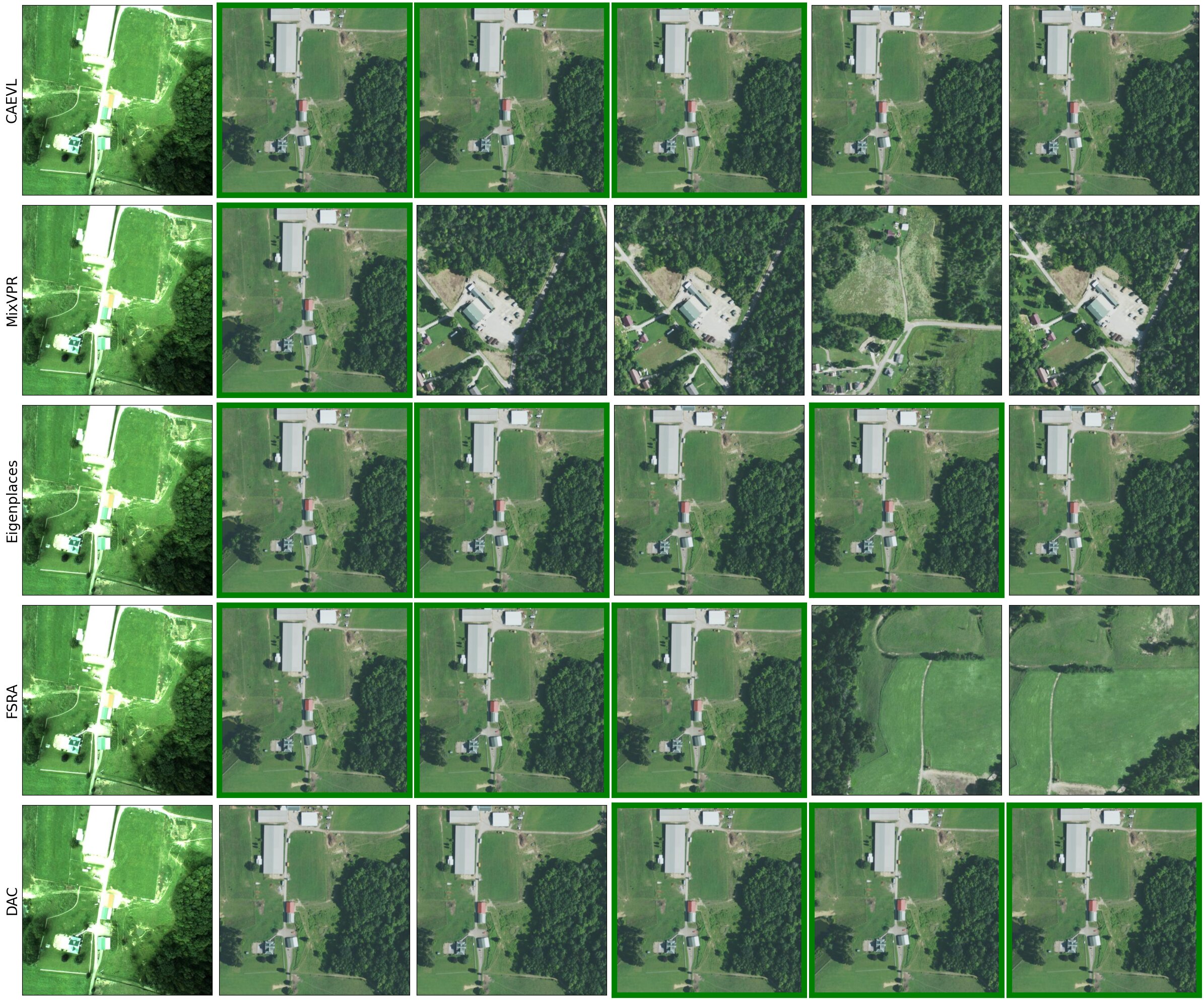}
    \caption{Visual comparison of top-5 retrieval results on a query from the ALTO dataset. For each method, the five best matches are shown to the right of a randomly picked query image. Correct predictions, defined as those within 15m of the ground truth, are highlighted with a green frame.}
    \label{fig: alto__query_2}
\end{figure*}

\begin{figure*}
    \centering
    \includegraphics[width=1\linewidth]{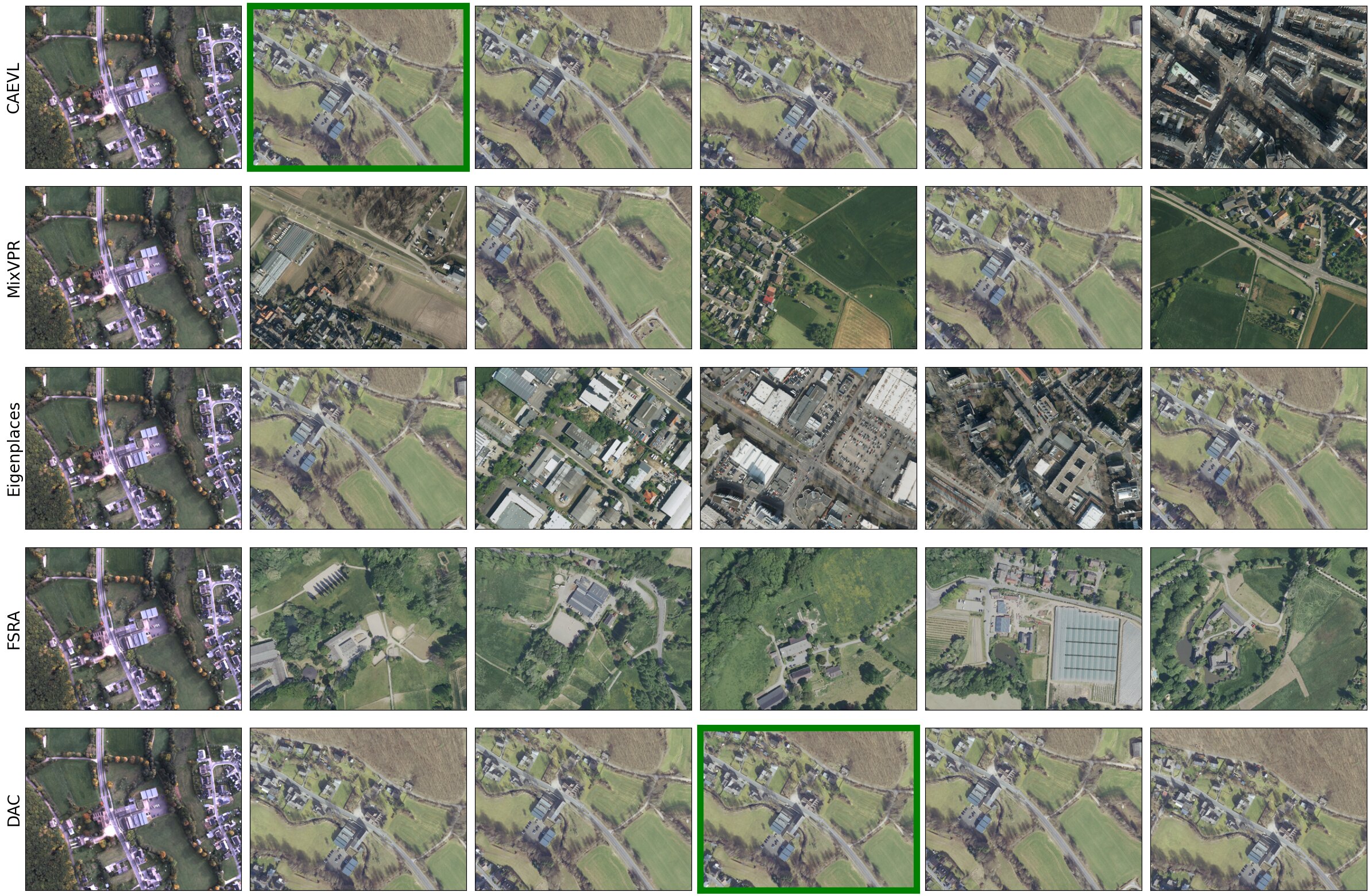}
    \caption{Visual comparison of top-5 retrieval results on a query from the VPAIR dataset. For each method, the five best matches are shown to the right of a randomly picked query image. Correct predictions, defined as those within 15m of the ground truth, are highlighted with a green frame.}
    \label{fig: vpair__query_316}
\end{figure*}

\begin{figure*}
    \centering
    \includegraphics[width=1\linewidth]{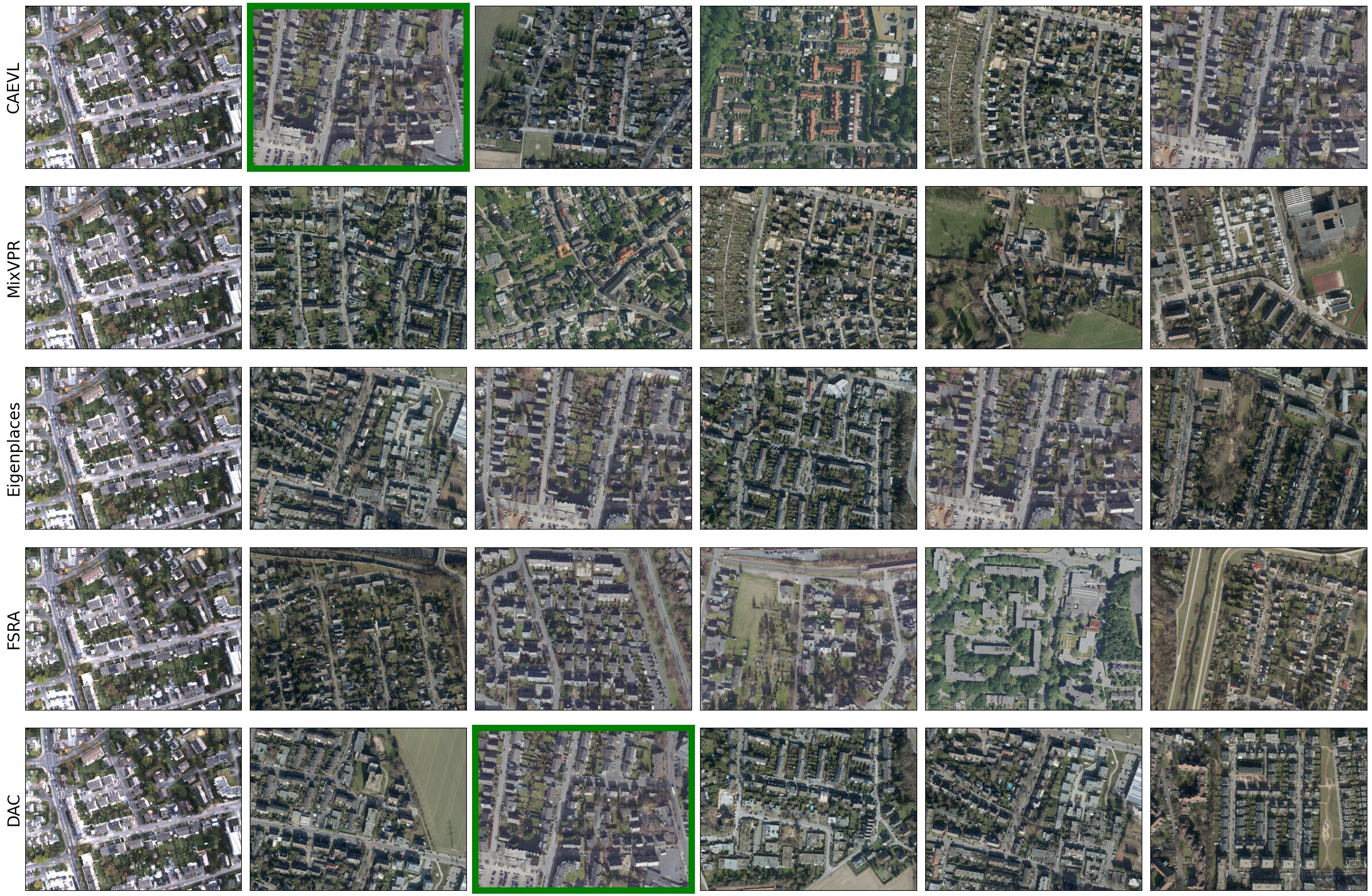}
    \caption{Visual comparison of top-5 retrieval results on a query from the VPAIR dataset. For each method, the five best matches are shown to the right of a randomly picked query image. Correct predictions, defined as those within 15m of the ground truth, are highlighted with a green frame.}
    \label{fig: vpair__query_552}
\end{figure*}

\begin{figure*}
    \centering
    \includegraphics[width=1\linewidth]{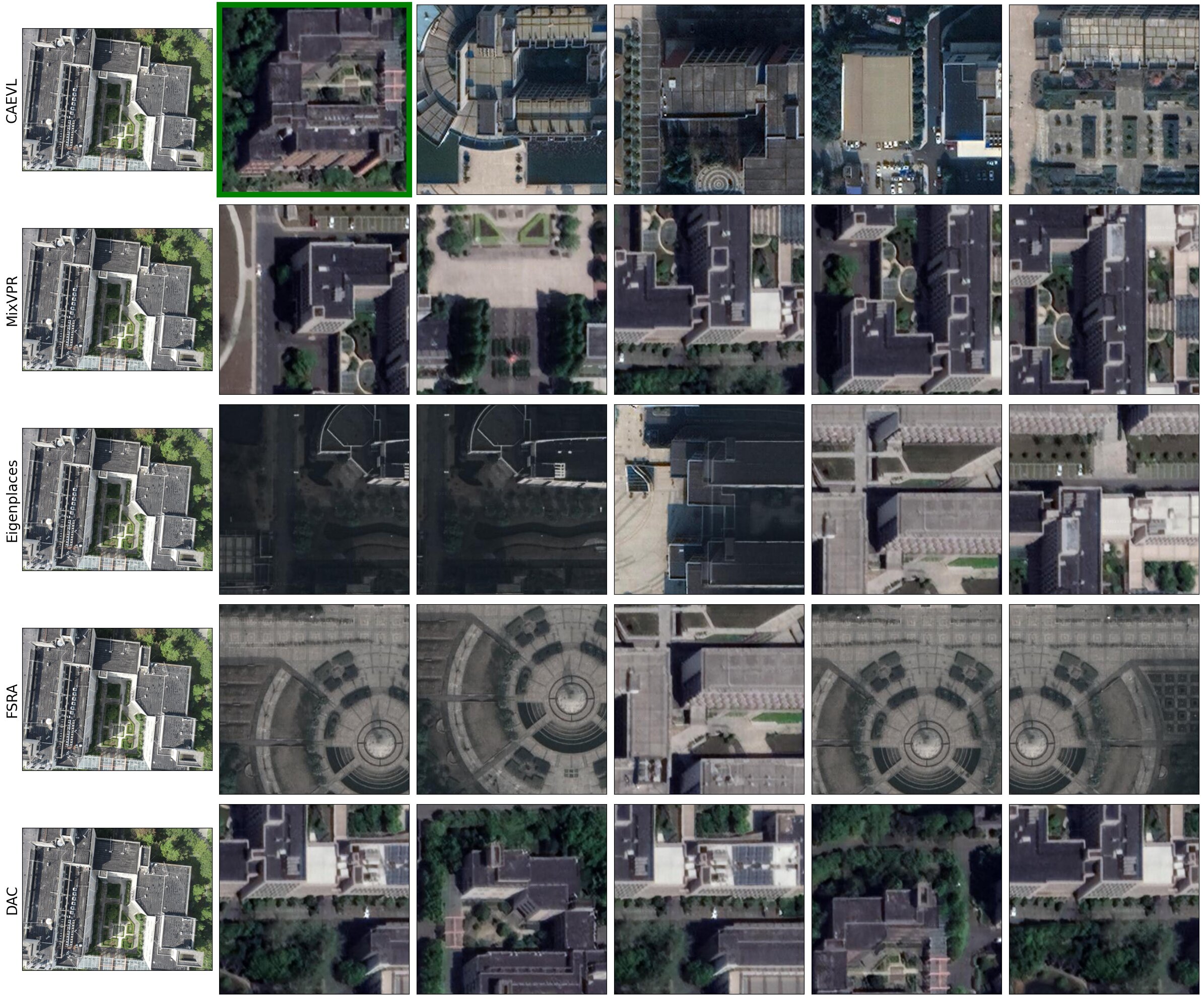}
    \caption{Visual comparison of top-5 retrieval results on a query from the DenseUAV dataset. For each method, the five best matches are shown to the right of a randomly picked query image. Correct predictions, defined as those within 15m of the ground truth, are highlighted with a green frame.}
    \label{fig: denseuav__query_7}
\end{figure*}

\begin{figure*}
    \centering
    \includegraphics[width=1\linewidth]{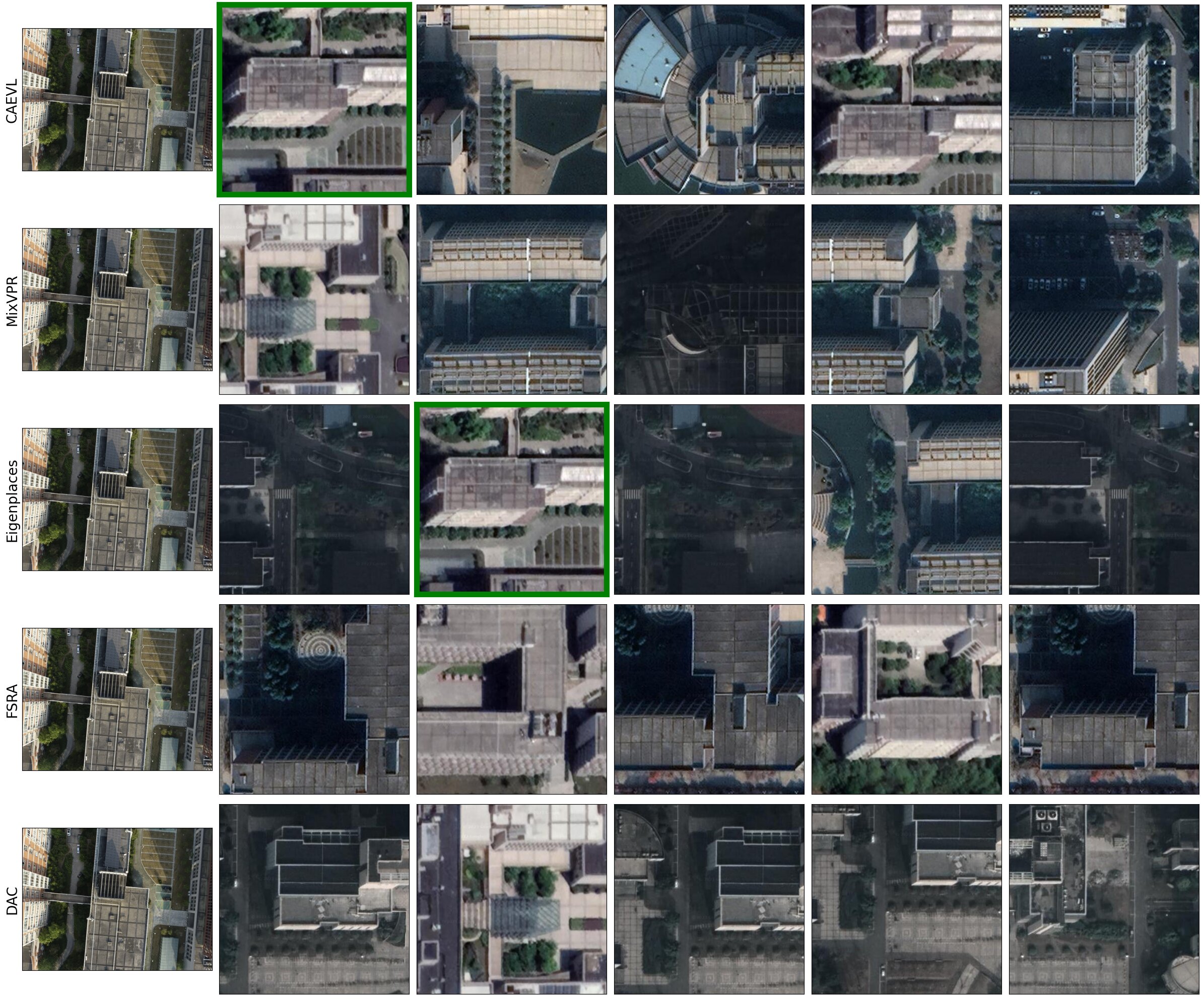}
    \caption{Visual comparison of top-5 retrieval results on a query from the DenseUAV dataset. For each method, the five best matches are shown to the right of a randomly picked query image. Correct predictions, defined as those within 15m of the ground truth, are highlighted with a green frame.}
    \label{fig: denseuav__query_124}
\end{figure*}

Figures \ref{fig: query_7261}, \ref{fig: query_9071} and \ref{fig: query_24501} provide a qualitative comparison of the top-5 predictions for our method, \method, against key baselines: MixVPR, EigenPlaces, FSRA, and DAC on the \dataset~dataset. Furthermore, Figures \ref{fig: alto__query_2}, \ref{fig: vpair__query_316}, \ref{fig: vpair__query_552}, \ref{fig: denseuav__query_7} and \ref{fig: denseuav__query_124} provide qualitative comparison examples on other datasets, namely the ALTO, VPAIR and DenseUAV datasets.

\end{document}